\documentclass{article}

\usepackage{iclr2027_conference,times}
\iclrfinalcopy

\usepackage[utf8]{inputenc} 
\usepackage[T1]{fontenc}    
\usepackage{hyperref}       
\usepackage{url}            
\usepackage{booktabs}       
\usepackage{amsfonts}       
\usepackage{nicefrac}       
\usepackage{microtype}      
\usepackage[table]{xcolor}  
\usepackage[dvipsnames]{xcolor}

\usepackage{kotex}
\usepackage{subcaption}

\usepackage{algorithm}
\usepackage{algorithmic}
\usepackage{amsmath}
\usepackage{amssymb}
\usepackage{mathtools}
\usepackage{amsthm}
\usepackage{array}
\usepackage{graphicx}
\usepackage{tabularx}
\usepackage{wrapfig}

\usepackage{multirow}
\usepackage{rotating}

\usepackage{makecell}
\usepackage{dblfloatfix}

\usepackage{verbatim}
\usepackage{fancyvrb}
\usepackage{upquote}

\definecolor{Gray}{gray}{0.8}
\usepackage{lineno}

\newcommand{\Reft}{\textsc{REFT}}
\newcommand{\thinkmark}{\texttt{<think>}}
\usepackage{tcolorbox}
\tcbuselibrary{breakable, skins}

\newtcolorbox{promptbox}[1][]{%
    enhanced,
    breakable,
    colback=gray!5,
    colframe=gray!50,
    boxrule=0.4pt,
    arc=2pt,
    left=6pt, right=6pt, top=4pt, bottom=4pt,
    fonttitle=\bfseries\small,
    coltitle=black,
    fontupper=\small,
    attach boxed title to top left={xshift=6pt, yshift=-2pt},
    boxed title style={
        colback=blue!10,
        colframe=gray!50,
        boxrule=0.4pt,
        arc=1pt,
    },
    #1
}

\newtcolorbox{promptbox2}[1][]{enhanced, breakable,
    colback=Thistle!5, colframe=gray!60, boxrule=0.4pt, arc=2pt,
    left=6pt, right=6pt, top=4pt, bottom=4pt,
    fonttitle=\bfseries\small, coltitle=black, fontupper=\small,
    attach boxed title to top left={xshift=3pt, yshift=-4pt},
    boxed title style={colback=Thistle!5, colframe=gray!60,
        boxrule=0.4pt, arc=1pt}, #1}

\newtcolorbox{rolloutbox}[1][]{%
    enhanced,
    breakable,
    colback=white,
    colframe=gray!50,
    boxrule=0.3pt,
    arc=1pt,
    left=5pt,
    right=5pt,
    top=3pt,
    bottom=3pt,
    fonttitle=\bfseries\footnotesize,
    coltitle=black,
    fontupper=\footnotesize,
    minipage boxed title*=-25pt,
    attach boxed title to top left={
        xshift=3pt,
        yshift=-2pt
    },
    boxed title style={
        colback=SeaGreen!10,
        colframe=gray!40,
        boxrule=0.3pt,
        arc=1pt
    },
    #1
}

\usepackage[capitalize,noabbrev]{cleveref}
\crefname{figure}{Figure}{Figures}
\crefname{table}{Table}{Tables}
\crefname{equation}{Equation}{Equations}
\AtBeginDocument{\renewcommand{\cref}[1]{\Cref{#1}}}

\newcommand{\thinktok}{\texttt{<think>}}
\newcommand{\reft}{REFT}

\definecolor{citecolor}{HTML}{0071BC}
\hypersetup{
    colorlinks=true,
    citecolor=citecolor
}

\title{Diversifying RLVR Rollouts via First-Token Exploration}

\author{%
\normalfont
\raisebox{-4pt}[\height][\depth]{%
\makebox[\dimexpr\textwidth-2\tabcolsep\relax][c]{%
    \begin{tabular}{c}
        \textbf{Soeun Kim}
        \hspace{4em}
        \textbf{Albert No}\thanks{
            Corresponding author: Albert No~\texttt{<albertno@yonsei.ac.kr>}
        }
        \\[8pt]
        Department of Artificial Intelligence, Yonsei University
    \end{tabular}%
}%
}%
}

\begin{document}

\maketitle

\begin{abstract}
Reinforcement learning with verifiable rewards (RLVR) trains reasoning models without labeled trajectories, using groups of verifier-scored rollouts to explore alternative reasoning paths.
Limited rollout diversity is a central bottleneck, typically addressed through adjustments to temperature, prefixes, or rollout selection.
We identify the \textit{first token} of the response as a structurally distinct target for diversification, largely overlooked in prior work.
We find that the first-token distribution is sharply concentrated and only weakly related to downstream correctness, as lower-probability candidates can yield similarly accurate responses.
Diversifying the first token can therefore broaden the reasoning paths explored within each rollout group with little loss in response quality.
Motivated by this observation, we introduce REFT (Rollout Exploration with First-Token Diversification), a lightweight modification to RLVR.
REFT samples first tokens uniformly from the policy's top-$N$ candidates and allocates rollouts evenly across the sampled tokens, leaving the rest of the pipeline unchanged.
We evaluate REFT on eight models spanning multiple architectures and sizes (0.5B--14B), with mathematical reasoning and code-generation tasks under GRPO and DAPO.
Across these settings, REFT consistently improves Pass@1, Pass@8, and Pass@64.
It also outperforms competing diversification methods at every evaluated budget, incurring the lowest rollout cost.

\end{abstract}

\section{Introduction}

Reinforcement learning with verifiable rewards (RLVR) has become the standard recipe for training reasoning models~\citep{guo2025deepseek,huopen}. 
Instead of imitating annotated solutions, the policy samples candidate reasoning traces, an automatic verifier scores their final answers, and the model is updated from the resulting rewards.  
In grouped methods such as GRPO~\citep{shao2024deepseekmath} and DAPO~\citep{yudapo}, this update is constructed from a finite set of rollouts for the same prompt.  
The rollout group is therefore the learner's window onto the space of possible reasoning paths.
If all rollouts receive the same reward, the group provides little contrastive signal.
If the rollouts are semantically redundant, the verifier can only compare a narrow set of alternatives.

This makes rollout diversity a central bottleneck in RLVR.    
A prominent line of work looks for useful variation where the model already hesitates, at high-entropy reasoning pivots~\citep{wangbeyond,yu2026erpo,wei2026entropy}, trajectory-level branches~\citep{xing2025lookahead,hou2025treerl,zhao2026reinforced,hu2025diversity,wan2026dsdr}, or final-answer alternatives~\citep{song2025outcome,wu2025invisible,zhusurprising}.
Other methods broaden exploration through temperature-controlled sampling~\citep{zhuang2025exploring,yang2025let,liu2025explore}, while efficiency-oriented methods reduce rollout cost through reuse and selective sampling~\citep{huang2026pros,liu2025spec,chang2026srt,chen2026jackpot}.
These works differ in mechanism, but they largely leave low-entropy prefix positions comparatively overlooked as potential sources of useful variation.

We challenge this assumption by focusing on the \textit{first token} of the response.
In mathematical traces, this is usually a discourse opener such as ``First'', ``Let'', or ``To'', rather than a substantive reasoning step, so it
appears too minor to matter.
We show the opposite. 
The first token is low-load in semantic content, but high-leverage in distributional effect. 
Because it is the first autoregressive choice in the reasoning trace, changing it alters the conditional distribution over every subsequent token.
Crucially, low probability at this position need not imply a poor chance of completing the problem correctly.
Our diagnosis reveals a large mismatch between first-token probability and downstream correctness, where the first-token distribution is sharply biased, yet rollout correctness remains nearly flat across the top-20 alternatives on Qwen, Llama, and Gemma models from 3B to 72B.
Moreover, this routing effect is not confined to the opener itself.  
After stripping the first token, continuations generated from different first tokens occupy different semantic regions, even though all subsequent tokens are sampled normally with the unchanged decoder, and uniform top-20 first-token sampling produces substantially higher diversity than standard sampling.
Therefore, a sharply concentrated first-token prior can leave viable regions of the continuation space underexplored.

These observations motivate a different remedy to spend a fixed rollout budget, \textbf{\Reft{}}: Rollout Exploration with First-Token Diversification.
For each prompt, \Reft{} takes the policy's own top-$N$ valid first-token candidates, samples $K$ of them uniformly, and allocates the rollout budget evenly across the selected first tokens.  
Beyond this first-token allocation, continuation decoding, verification, reward and advantage computation, the RL objective, and the total rollout budget are unchanged. 
\Reft{} diversifies where the model is artificially certain and correctness is comparatively stable, rather than injecting noise where the model is uncertain and correctness is fragile.

Empirically, \Reft{} improves Pass@1, Pass@8, and Pass@64 under both GRPO and DAPO on mathematical reasoning, across six models spanning 0.5B--14B parameters.
The gains also extend to code-generation RLVR, and a matched comparison shows REFT outperforming prior diversification alternatives at every budget while using the lowest rollout cost tested.
A first-token loss-masking control attributes the gains to broader rollout coverage rather than to corrected first-token crediting.
Together, these results suggest a simple inversion of the usual exploration intuition. 
In RLVR, useful diversity can be recovered not only from high-entropy reasoning forks, but also from low-load first-token choices whose probability distribution is sharply biased yet only weakly tied to correctness.

\section{Preliminaries}
\label{sec:prelim}

\paragraph{RLVR and grouped rollouts.}
Reinforcement learning with verifiable rewards (RLVR) trains a language-model policy from sampled attempts rather than labeled reasoning trajectories.  Given a prompt $x$, the policy samples a \emph{rollout} $y$, comprising reasoning trace and the final answer, and an automatic verifier returns a scalar reward $R(x,y)$. 
Group-based RLVR methods sample $G$ rollouts for the same prompt and update the policy from their relative rewards.  
In GRPO~\citep{shao2024deepseekmath}, the reward for rollout $y_i$ is converted to a group-normalized advantage
\begin{equation}
    A_i =
    \frac{R_i - \operatorname{mean}(\{R_j\}_{j=1}^{G})}
    {\operatorname{std}(\{R_j\}_{j=1}^{G}) + \epsilon},
    \label{eq:grpo_advantage}
\end{equation}
where $R_i=R(x,y_i)$, and a PPO-style clipped objective \citep{schulman2017proximal} then applies this trajectory-level advantage to the tokens of $y_i$.
Therefore, the finite rollout group is the unit from which the learning signal is constructed and recent RLVR systems and variants build on this template. 
DeepSeek-R1~\citep{guo2025deepseek} scales verifier-based reasoning RL, and DAPO~\citep{yudapo} improves GRPO with dynamic sampling, asymmetric clipping, token-level loss, and overlong-response handling.
Recent methods further improve grouped-rollout data construction or usage.
GRESO~\citep{zhengact} filters uninformative prompts, Jackpot~\citep{chen2026jackpot} budgets rejection sampling, AR3PO~\citep{zhang2025improving} reuses rollouts, RL-ZVP~\citep{le2025no} learns from zero-variance prompts, PODS~\citep{xu2025not} selects informative rollout subsets for update, and saturated all-correct prompts are reheated to extract residual signal \citep{liu2025explore}.

\paragraph{Why rollout diversity matters.}
Equation~\eqref{eq:grpo_advantage} makes diversity a first-order concern.   
If all rollouts in a group receive the same reward, the group has no reward contrast and contributes little or no policy-gradient signal.
Even when rewards are non-identical, a group of near-duplicate reasoning traces provides a narrow signal, because the verifier can only compare alternatives that were actually sampled.
Diverse rollouts therefore serve two roles.  
They increase the probability that a finite group contains useful reward variation, and they broaden the set of reasoning regions over which the verifier can assign credit.
Under Pass@\(k\) evaluation, retaining multiple plausible solution modes can matter as much as increasing the probability of the model response. 
Recent analyses of outcome-level diversity collapse~\citep{song2025outcome}, support narrowing under RLVR~\citep{wu2025invisible}, and sample-reinforcement dynamics~\citep{zhusurprising} point to a common concern that improving Pass@1 can narrow the accessible rollout distribution and hurt exploration or inference-time scaling.

\paragraph{Existing routes to rollout diversity.}
Maintaining rollout diversity has accordingly drawn a wide range of remedies. 
The most immediate is adjusting the sampling temperature, varied across rollouts, applied selectively to high-entropy tokens \citep{zhuang2025exploring}, or annealed from hot to cold along the sequence \citep{yang2025let}.
Much recent work concentrates exploration where the model hesitates. 
Policy gradients are reserved for the high-entropy minority of ``forking'' tokens \citep{wangbeyond}, ERPO amplifies advantages at high-entropy decision pivots \citep{yu2026erpo},
and rollout generation grows into trees, with Entropy-Tree and LATR branching where uncertainty runs high \citep{wei2026entropy, xing2025lookahead}, FR3E relaunching from high-entropy decision points \citep{zheng2025first}, and TreeRL searching on-policy \citep{hou2025treerl}.
Others reward diversity directly, as an incentive in the objective \citep{hu2025diversity, wan2026dsdr}, as bonuses over final answers \citep{song2025outcome}, or as a set-level Pass@$k$ training target \citep{chen2025pass}. 
Different as they are, these approaches share one instinct, how the response's first token is handled. 
It is treated as fixed scaffolding, amortized as a shared prefix \citep{huang2026pros, chang2026srt}, reused speculatively across iterations \citep{liu2025spec}, or pinned at temperature 1.0 by annealed decoding to preserve template-like text \citep{yang2025let}. 
Our question is complementary: can useful alternatives remain hidden at the very first choice that the model sees \textit{no question to ask} and existing approaches largely \textit{take for granted}?


\section{Diagnosis: Low-Load, High-Leverage First Tokens} 
\label{sec:observations}

We define the \emph{first token} as the earliest semantic token of the response, after leading whitespace and any control markers such as \thinkmark{}. 
In math traces it is typically a discourse opener such as ``First'', ``Let'', or ``To'' (Appendix~\ref{app:per_prompt_first_token}) rather than an algebraic operation, a case distinction, or an answer.
Given its limited semantic content, it may seem an unlikely place to find meaningful variation.
This section shows that useful diversity can begin with this seemingly minor choice.
\begin{figure*}[!h]
    \vspace{-4pt}
    \centering
    \includegraphics[width=\linewidth]
    {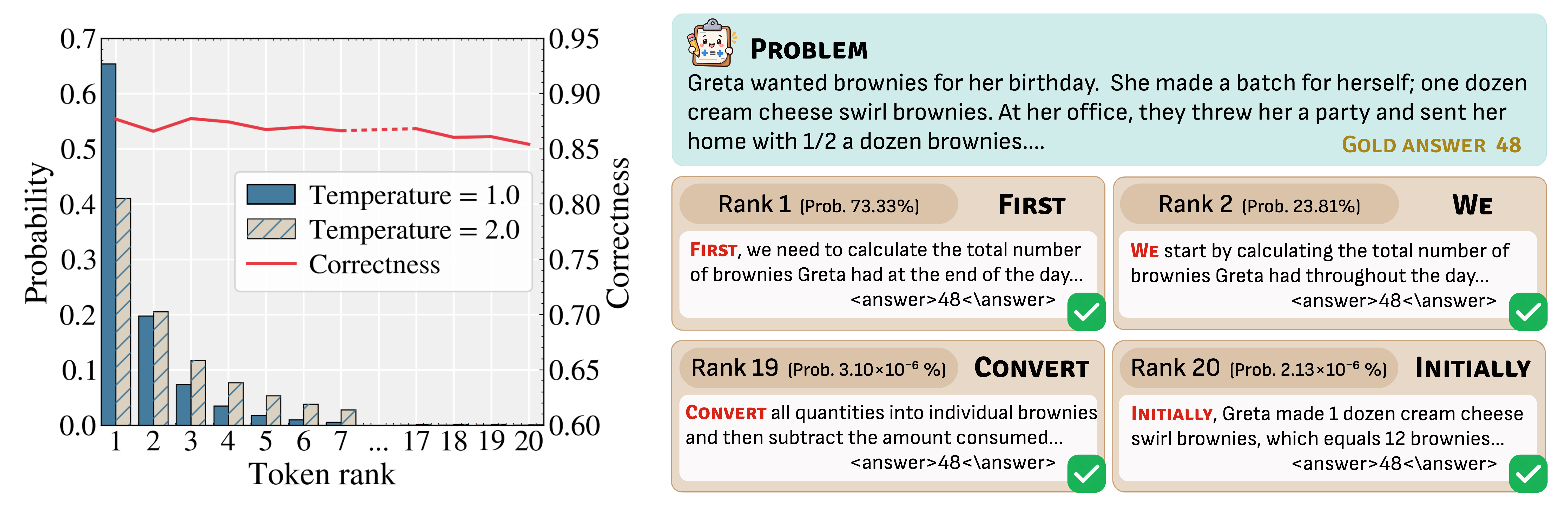}
    \par\vspace{-7pt}
    \begin{subfigure}[t]{0.44\linewidth}
        \centering
        \label{fig:prob_correctness}
    \end{subfigure}%
    \begin{subfigure}[t]{0.56\linewidth}
        \centering
        \label{fig:opener_examples}
    \end{subfigure}
    \par\vspace{-4pt} 
    \caption{
    \textbf{Sharp probability, flat correctness.}
    (Left) The model assigns high probability to the top first token, but rollout correctness remains nearly flat across the top-20 ranks~(Qwen2.5-7B-Instruct, GSM8K).
    (Right) Despite their low sampling probabilities, Convert (rank-19) and Initially (rank-20) produce natural continuations that reach the same correct answer as the rank-1 and rank-2 first tokens.
    }
    \label{fig:main_fig}
    \vspace{-14pt}
\end{figure*}

\paragraph{Low-probability first tokens perform reasonably well.}\label{subsec:first_token_decoupling}

For each prompt $x$, write $c(x)$ for everything the model conditions on before its first token, the prompt, template prefill, and any skipped markers (instantiated per model in Appendix~\ref{app:cx}), and let $f_r(x)$ be the rank-$r$ token of the first-token distribution $\pi_\theta(\cdot\mid c(x))$.
The model's prior over this position is extremely sharp.  
Figure~\ref{fig:main_fig}~(Left) shows that the top-ranked first token has mean probability $0.653$ at temperature $1.0$, versus $1.30\times10^{-5}$ for rank 20. 
The surprise is how little correctness changes as probability falls so sharply across ranks.
We fix the first token to be $f_r(x)$ and let the unchanged
decoder complete the rollout.
Correctness remains nearly flat across the top-20 ranks,
with rank 20 achieving $85.41\%$ against rank 1's $87.69\%$,
despite a roughly $50{,}000$-fold difference in mean sampling
probability (Figure~\ref{fig:main_fig}~(Left); red line).

This is the central empirical asymmetry. 
At the first-token position, model probability and verifier correctness decouple.  
Standard sampling commits the group to a handful of openers before any substantive reasoning exists, yet a low-probability first token is not a bad reasoning choice.
Up to 72B and recent models, the first-token distribution is strikingly concentrated despite only modest differences in correctness across alternative openings (Appendix~\ref{app:extend_distributions}).
Rank-1/rank-20 probability ratios span roughly $10^3$--$10^7$, without a commensurate separation in downstream correctness.

\paragraph{The first token influences the whole continuation.}
\label{subsec:first_token_routing}

\begin{wrapfigure}{r}{0.385\textwidth}
    \centering
    \vspace{-0.9em}
    \includegraphics[width=0.38\textwidth]{
    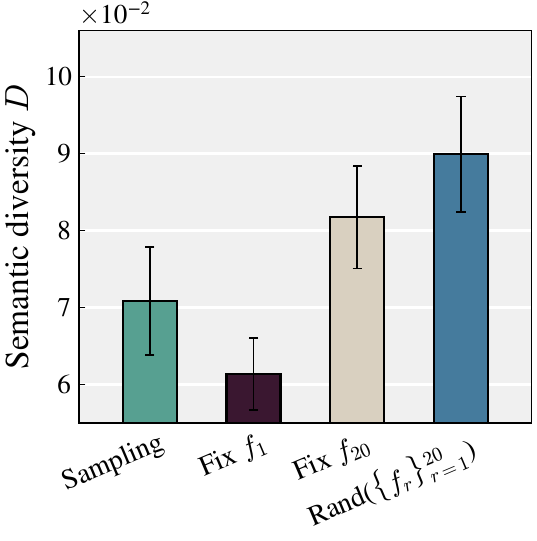}
    \vspace{-2.5em}
    \caption{
    \small
    \textbf{First tokens route continuations.}
    Semantic diversity is measured after stripping the first token itself, so the effect comes from the rest of the rollout.
    }
    \label{fig:semantic_diversity}
    \vspace{-1.0em}
\end{wrapfigure}

The first token matters not only for what it says locally, but for how it conditions the rest of the rollout.
Decomposing a response into the first token $F$ and continuation $z$,
\begin{equation}
\pi_\theta(F, z \mid c(x)) \;=\; \pi_\theta(F \mid c(x))\,\pi_\theta(z \mid c(x), F),
\label{eq:routing}
\end{equation}
varying $F$ induces distinct continuation distributions, and even a token with tiny marginal probability can open a competent but rarely visited region of solutions.

To test whether this routing effect is real, we compare four rollout strategies at matched group size: standard sampling, fixing $f_1(x)$, fixing $f_{20}(x)$, and sampling uniformly from the top-20 first tokens.
We strip the first token from each rollout and measure the semantic diversity of what remains (Appendix~\ref{app:semantic_diversity}).
Figure~\ref{fig:semantic_diversity} shows that first-token choice changes the continuation distribution. 
Because diversity is measured after stripping, the effect is not surface variation in the first token. 
Even a fixed rank-20 opener yields more diverse continuations than standard sampling, and uniform sampling over the top-20 yields the most, suggesting that different first tokens contribute complementary rather than redundant variety.
Comparable correctness and distinct continuations together mean the sharp prior is hiding \emph{coverage}, solutions the group would benefit from contrasting yet standard sampling almost never visits.

\paragraph{Training can further sharpen the first-token prior.} \label{subsec:grpo_first_token_credit}
\begin{wrapfigure}{r}{0.385\textwidth}
    \vspace{-1.8em}
    \centering
    \includegraphics[width=0.38\textwidth]{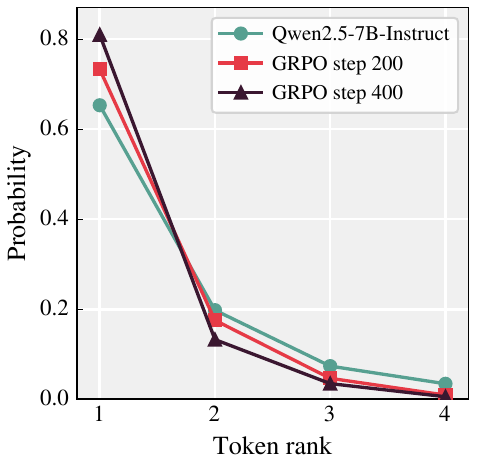}
    \vspace{-2.0em}
    \caption{
    \small
    \textbf{RLVR sharpens the first-token prior.}
    The top-1 first-token probability increases during training, although the first token is only weakly tied to correctness.
    }
    \label{fig:main_rl_sharpener}
    \vspace{-0.2em}
\end{wrapfigure}

The bias is not only present before RLVR training; GRPO-style training can amplify it.  
The trajectory-level advantage is applied to every token in the rollout.  
For a rollout starting with $F_i$, the first-token gradient term is proportional to
\begin{equation*}
    A_i \nabla_\theta \log \pi_\theta(F_i \mid c(x)),
\end{equation*}
where $A_i$ is the group-normalized advantage from Equation~\eqref{eq:grpo_advantage}. 
This assigns the same scalar credit to the generic opener as to the later tokens that actually solve the problem.
Because the top-ranked first token is sampled far more often, it also has far more chances to co-occur with positive-advantage trajectories.
As Figure~\ref{fig:main_rl_sharpener} shows, the top-1 first-token probability climbs steadily during training.
The first-token sharpening we observe is consistent with prior work identifying entropy collapse as a barrier to exploration in RLVR~\citep{yudapo,zheng2025first,liu2026prorl,cui2025entropy}.
By further concentrating the first-token distribution, standard GRPO can restrict access to diverse continuations before substantive reasoning begins.

\begin{wrapfigure}{r}{0.6\linewidth}
\vspace{-1.0em}
\centering
\begin{subfigure}[t]{0.49\linewidth}
    \centering
    \includegraphics[width=\linewidth]{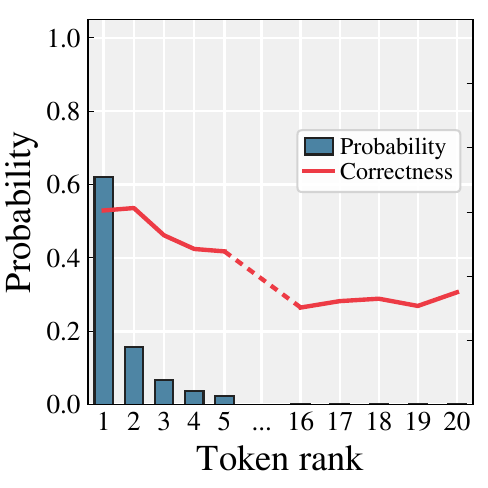}
    \caption{Base}
    \label{fig:gemma_base}
\end{subfigure}
\hspace{-0.017\linewidth}
\begin{subfigure}[t]{0.49\linewidth}
    \centering
    \includegraphics[width=\linewidth]{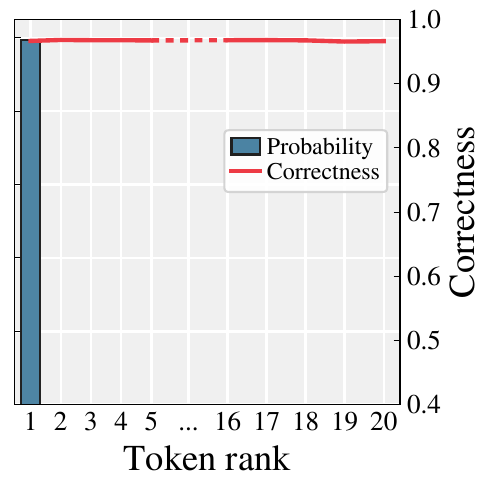}
    \caption{Instruct}
    \label{fig:gemma_it}
\end{subfigure}
\vspace{-6pt}
\caption{
\small
\textbf{A sharper first-token prior after post-training.}
Gemma4-31B on GSM8K.
The instruct model concentrates on rank 1, unlike the relatively broader, rank-sensitive base model.
}
\label{fig:gemma}
\vspace{-27pt}
\end{wrapfigure}

The Gemma4-31B~\citep{team2026gemma} base--instruct comparison in Figure~\ref{fig:gemma} provides an informative contrast.
The base prior is substantially softer and its correctness declines across ranks, whereas the instruct prior collapses onto a single token even though correctness flattens. 
While this comparison cannot attribute the sharpening specifically to RLVR, it suggests that post-training can introduce or amplify a strong preference for particular first tokens without a commensurate separation in downstream correctness.
Full base--instruct curves through rank 20 are in Appendix~\ref{app:extend_distributions} (Figure~\ref{fig:gemma_base_instruct_full}).

\paragraph{Why temperature is not the answer.}
\label{subsec:temperature_not_enough}
A natural remedy for increasing rollout diversity is to raise the sampling temperature.
However, the prior remains too sharp for temperature to fix at acceptable cost.
As shown in Figure~\ref{fig:main_fig}, even $T{=}2.0$ leaves the rank-20 candidate with probability ${\sim}1.8\times10^{-3}$, and with group size $G{=}8$, the probability of sampling it at least once is ${\sim}1.4\%$.
Meanwhile, every later conditional along the trajectory, including fragile mid-reasoning steps, is flattened as collateral. 
So, what we need is not a global increase in sampling temperature, but targeted coverage of plausible first-token alternatives. 
We also compare \reft{} against temperature-based alternatives in Section~\ref{sec:headtohead}.


\section{\Reft{}: Rollout Exploration with First-Token Diversification}
\label{sec:method}
Section~\ref{sec:observations} identifies the first token as a promising target for exploration.
It is \textit{low-load}: diversifying across lower-ranked first tokens incurs little loss in correctness.
It is \emph{high-leverage}: lower-probability candidates are rarely sampled, yet changing the first token can affect every subsequent conditional distribution.
Together, these make the first token a surprisingly efficient exploration site, one that the sharp prior leaves underexplored under standard sampling.

These observations motivate \textbf{REFT} (Rollout Exploration with First-Token Diversification), which changes only first-token sampling.
For each prompt, \reft{} samples first tokens uniformly from the policy's top candidates.
This encourages each rollout group to explore a broader range of starting tokens rather than concentrate on the most probable ones.
Subsequent tokens are generated as usual.

\paragraph{Candidate set.}
Given a prompt $x$, we compute the policy distribution at the first-token position and keep the top-$N$ valid candidates:
\begin{equation*}
\mathcal{F}_N(x) = \{f_1(x), \ldots, f_N(x)\}, \qquad \pi_{\theta_{\mathrm{old}}}(f_1 \mid c(x)) \geq \cdots \geq \pi_{\theta_{\mathrm{old}}}(f_N \mid c(x)).
\end{equation*} 
Here $f_1,\ldots,f_N$ are the highest-probability valid first tokens under the behavior policy. 
Because the candidates come from the policy's own top-$N$ set, \Reft{} does not introduce external prompts or hand-written prefixes.
It reallocates budget among first tokens the model already considers plausible.

\paragraph{First-token diversification.}
Standard sampling draws the first token of each of the $G$ rollouts from the policy distribution.
A simple alternative samples $G$ first tokens uniformly from $\mathcal{F}_N(x)$.
However, exploring many first tokens leaves few continuations per token, making each token's expected return difficult to estimate reliably.
To balance coverage and repeated sampling, \reft{} selects $K$ first tokens uniformly without replacement, with $K$ dividing $G$:
\begin{equation*}
    \mathcal{S}_K(x)
    \sim \mathrm{Unif} \left( \left\{ \mathcal{S}\subseteq \mathcal{F}_N(x): |\mathcal{S}|=K \right\} \right).
\end{equation*}
For each token in $\mathcal{S}_K(x)$, we generate $G/K$ independent continuations using the standard sampling procedure, yielding $G$ rollouts in total:
\begin{equation*}
    z_{f,j}
    \sim \pi_{\theta_{\mathrm{old}}} \left( \cdot \mid c(x),f \right), \qquad j=1,\ldots,G/K.
\end{equation*}
Smaller $K$ allows more continuations per selected token, while $K=G$ gives one per token and recovers the simple alternative above.
The resulting rollout group is
\begin{equation*}
    \mathcal{Y}^{\mathrm{REFT}}_G(x)
    = \left\{ (f,z_{f,j}) : f\in\mathcal{S}_K(x), \; j=1,\ldots,G/K \right\}.
\end{equation*}
Therefore, \Reft{} replaces sampling from a sharply peaked first-token distribution with explicit coverage of $K$ plausible first tokens, while leaving all subsequent generation unchanged.

\vspace{-4pt}
\begin{algorithm*}[!h]
\caption{\Reft{} rollout sampler}
\label{alg:reft}
\begin{algorithmic}[1]
\REQUIRE Prompt $x$ with response-start context $c(x)$, behavior policy $\pi_{\theta_{\mathrm{old}}}$, group size $G$, candidate width $N$, selected-token count $K$
\ENSURE Rollout group $\mathcal{Y}^{\mathrm{REFT}}_G(x)$

\STATE $\mathcal{F}_N(x) \gets \{f_1(x),\ldots,f_N(x)\}$ 
\hfill\COMMENT{Top-$N$ valid tokens under $\pi_{\theta_{\mathrm{old}}}(\cdot \mid c(x))$}
\STATE $\mathcal{S}_K(x) \sim \mathrm{Unif}\!\left( \left\{ \mathcal{S}\subseteq\mathcal{F}_N(x): |\mathcal{S}|=K \right\} \right)$
\STATE For each $f\in\mathcal{S}_K(x)$, sample
$ z_{f,1},\ldots,z_{f,G/K}
\sim \pi_{\theta_{\mathrm{old}}} \left( \cdot \mid c(x), f \right)$

\STATE $\mathcal{Y}^{\mathrm{REFT}}_G(x)\gets
\left\{ (f,z_{f,j}) : f\in\mathcal{S}_K(x),\; j=1,\ldots, G/K\right\}$

\RETURN $\mathcal{Y}^{\mathrm{REFT}}_G(x)$
\end{algorithmic}
\end{algorithm*}

\paragraph{What remains unchanged.}
After constructing $\mathcal{Y}^{\mathrm{REFT}}_G(x)$, optimization proceeds exactly as in the underlying RLVR algorithm.  
The verifier scores each rollout, group-normalized advantages are computed as usual, and the same GRPO/DAPO-style objective is applied to the sampled trajectories.  
\Reft{} acts entirely within rollout construction, as a drop-in modification for the sampler's first decision, and composes with existing refinements such as dynamic sampling and asymmetric clipping.

\paragraph{Rollout construction with a fixed RL loss.}
\Reft{} alters the data-collection distribution only at a single position, and the completed rollouts are trained with the same GRPO/DAPO objective as the baselines, without an importance correction for the first-token sampler.
The resulting update therefore reflects \Reft{}'s reallocation of the finite rollout budget across plausible first tokens, rather than an unbiased estimate of standard on-policy sampling.
Correcting back toward the original first-token prior would counteract the intended coverage.
Further discussion is in Appendix~\ref{app:off-policy}.

\section{Experiments} \label{sec:experiments}
\subsection{Experimental Setup} \label{sec:experimental_setup}
\begin{table*}[!t]
\centering
\caption{
\textbf{Mathematical reasoning across models and RLVR objectives.}
GRPO and DAPO are compared with their REFT variants.
Math~Avg.\ excludes GSM8K, and \textbf{bold} marks the higher score.
}
\setlength{\tabcolsep}{3.5pt}
\setlength{\aboverulesep}{2pt}
\setlength{\belowrulesep}{2pt}
\renewcommand{\arraystretch}{1.1}
\makebox[\textwidth][c]{%
\resizebox{1.04\textwidth}{!}{
\begin{tabular}{l!{\vrule width 0.6pt}cc!{\vrule width 0.6pt}cccccccc!{\vrule width 0.6pt}cc}
\toprule[1.1pt]
& \multicolumn{2}{c!{\vrule width 0.6pt}}{GSM8K}
& \multicolumn{2}{c}{MATH-500}
& \multicolumn{2}{c}{AIME24}
& \multicolumn{2}{c}{AIME25}
& \multicolumn{2}{c!{\vrule width 0.6pt}}{AMC23}
& \multicolumn{2}{c}{Math~Avg.} \\
\cmidrule(lr){2-3}
\cmidrule(lr){4-5}
\cmidrule(lr){6-7}
\cmidrule(lr){8-9}
\cmidrule(lr){10-11}
\cmidrule(lr){12-13}
Method
& Pass@1 & Pass@8
& Pass@1 & Pass@8
& Pass@1 & Pass@8
& Pass@1 & Pass@8
& Pass@1 & Pass@8
& Pass@1 & Pass@8 \\
\midrule[0.8pt]

\rowcolor{gray!10}
\multicolumn{1}{c!{\vrule width 0.6pt}}{\textbf{\textit{Qwen2.5-3B-Instruct}}}
& 80.14 & 94.16 & 51.40 & 78.00 & 6.67 & 16.67 & 0.00 & 13.33 & 35.00 & 75.00 & 23.27 & 45.75 \\
GRPO
& 84.53 & 95.38 & 59.80 & 79.40 & 10.00 & 20.00 & 3.33 & 16.67 & 42.50 & 77.50 & 28.91 & 48.39 \\
\rowcolor{cyan!7}
GRPO + REFT
& \textbf{86.28} & \textbf{96.13} & \textbf{61.00} & \textbf{81.20} & \textbf{13.33} & \textbf{23.33} & \textbf{6.67} & \textbf{20.00} & \textbf{47.50} & \textbf{85.00} & \textbf{32.13} & \textbf{52.38} \\
DAPO
& 86.05 & 95.15 & 60.00 & 78.60 & 10.00 & 20.00 & 6.67 & 16.67 & 45.00 & 80.00 & 30.42 & 48.82 \\
\rowcolor{cyan!7}
DAPO + REFT
& \textbf{86.20} & \textbf{96.06} & \textbf{61.20} & \textbf{80.00} & \textbf{13.33} & \textbf{26.67} & \textbf{13.33} & \textbf{20.00} & \textbf{52.50} & \textbf{85.00} & \textbf{35.09} & \textbf{52.92} \\
\specialrule{0.8pt}{2pt}{2pt}

\rowcolor{gray!10}
\multicolumn{1}{c!{\vrule width 0.6pt}}{\textbf{\textit{Llama3.2-3B-Instruct}}}
& 73.24 & 91.96 & 29.80 & 64.00 & 3.33 & 13.33 & 0.00 & 3.33 & 20.00 & 50.00 & 13.28 & 32.66 \\
GRPO
& 75.66 & 92.12 & 35.60 & 64.80 & 6.67 & 16.67 & 3.33 & 3.33 & 22.50 & 60.00 & 17.02 & 36.20 \\
\rowcolor{cyan!7}
GRPO + REFT
& \textbf{77.63} & \textbf{94.09} & \textbf{36.80} & \textbf{65.40} & 6.67 & \textbf{20.00} & 3.33 & \textbf{6.67} & \textbf{25.00} & \textbf{62.50} & \textbf{17.95} & \textbf{38.64} \\
DAPO
& 73.84 & 93.10 & 37.20 & 65.60 & 10.00 & 23.33 & 3.33 & 6.67 & 22.50 & 60.00 & 18.26 & 38.90 \\
\rowcolor{cyan!7}
DAPO + REFT
& \textbf{78.47} & \textbf{94.01} & \textbf{41.00} & \textbf{67.00} & \textbf{13.33} & 23.33 & \textbf{6.67} & \textbf{10.00} & \textbf{25.00} & \textbf{65.00} & \textbf{21.50} & \textbf{41.33} \\
\specialrule{0.8pt}{2pt}{2pt}

\rowcolor{gray!10}
\multicolumn{1}{c!{\vrule width 0.6pt}}{\textbf{\textit{Qwen2.5-7B-Instruct}}}
& 88.48 & 96.74 & 53.40 & 80.00 & 10.00 & 20.00 & 0.00 & 23.33 & 37.50 & 75.00 & 25.23 & 49.58 \\
GRPO
& 91.05 & 97.12 & 72.00 & 86.40 & 13.33 & 23.33 & 10.00 & 26.67 & 55.00 & 85.00 & 37.58 & 55.35 \\
\rowcolor{cyan!7}
GRPO + REFT
& \textbf{91.89} & \textbf{97.42} & \textbf{73.60} & \textbf{87.20} & \textbf{20.00} & \textbf{30.00} & \textbf{16.67} & \textbf{36.67} & \textbf{65.00} & 85.00 & \textbf{43.82} & \textbf{59.72} \\
DAPO
& 91.36 & 96.74 & 72.40 & 86.40 & 16.67 & 26.67 & 10.00 & 30.00 & 57.50 & 80.00 & 39.14 & 55.77 \\
\rowcolor{cyan!7}
DAPO + REFT
& \textbf{92.27} & \textbf{97.27} & \textbf{73.00} & \textbf{86.80} & \textbf{20.00} & \textbf{30.00} & \textbf{13.33} & \textbf{33.33} & \textbf{62.50} & \textbf{85.00} & \textbf{42.21} & \textbf{58.78} \\
\bottomrule[1.1pt]
\end{tabular}
}}
\label{tab:main_pass1_pass8}
\vspace{-12pt}
\end{table*}

\paragraph{Mathematical reasoning.}
We compare GRPO~\citep{shao2024deepseekmath} and DAPO~\citep{yudapo} with their \reft-augmented variants. 
Training covers six models spanning 0.5B to 14B: Qwen2.5-3B/7B-Instruct \citep{yang2024qwen25} and Llama3.2-3B-Instruct \citep{grattafiori2024llama} on GSM8K \citep{cobbe2021training}, BigMath-Easy, and BigMath \citep{albalak2025big}; Qwen2.5-0.5B-Instruct on GSM8K, all under both objectives; Qwen3.5-9B~\citep{qwen35blog} and Qwen2.5-14B-Instruct on BigMath with GRPO. 
GSM8K uses group size $G{=}8$, and BigMath subsets use $G{=}16$. 
GSM8K-trained models are evaluated on GSM8K; BigMath-trained models on MATH-500 \citep{lightman2023let}, AIME 2024/2025 \citep{aime24, aime25}, and AMC 2023 \citep{li2024numinamath}.
We report Pass@1, Pass@8, and Pass@64, with Math~Avg.\ the unweighted average of the four non-GSM8K benchmarks.

\paragraph{Code generation.}
DeepSeek-R1-Distill-Qwen-7B~\citep{guo2025deepseek} and Qwen2.5-Coder-3B-Instruct~\citep{hui2024qwen2} are trained with GRPO ($G{=}8$) on a decontaminated, test-verified Python set~\citep{openr1codeset}, using the fraction of passed sandbox unit tests as reward.
The former generates an explicit \thinktok{} trace, whereas the latter generates Python directly.
Evaluation reports Pass@1 and Pass@8 on HumanEval+, MBPP+~\citep{liu2023your}, and LiveCodeBench v4~\citep{jain2025livecodebench}.

\paragraph{Common configuration.}
Unless stated otherwise, \reft{} uses candidate width $N=20$, selected-token count $K=4$, allocating rollout slots evenly across selected first tokens.
Candidate-width and replication ablations are in
Appendix~\ref{app:ablation_stratify}.
Within each baseline--\reft{} comparison, all methods share the same rollout budgets and identical decoding configurations. 
Training rollouts are generated with vLLM~\citep{kwon2023efficient}.
Full training and evaluation configurations are in Appendix~\ref{app:exp_details}, and runtime measurements for the profiled configurations are reported in Appendix~\ref{app:runtime_overhead}.

\subsection{Mathematical Reasoning Across Scales and Generations}
\label{sec:mathresults}
\paragraph{Consistent gains across models and sampling budgets.}
Table~\ref{tab:main_pass1_pass8} reports Pass@1 and Pass@8 across three models, two objectives, and five benchmarks. 
\Reft{} consistently improves performance even on GSM8K, where the models start from strong baselines, and the gains grow larger on Math~Avg., showing that \Reft{} is especially effective when additional rollout exploration has more room to pay off (e.g., Qwen2.5-7B GRPO Pass@1: 37.58$\rightarrow$43.82).
The two metrics measure different things, Pass@1 the quality of a single sample and Pass@8 whether any of eight sampled responses is correct, so their simultaneous improvement shows that \reft{} benefits both individual samples and the rollout group as a whole. 
The gains persist across three training seeds~(Appendix~\ref{app:seeds}).

\begin{wraptable}{r}{0.55\linewidth}
\vspace{-12pt}
\centering
\caption{
\textbf{Scaling to larger and newer models.}
Pass@1 on BigMath-trained models with GRPO ($G{=}16$).
}
\vspace{-4pt}
\scriptsize
\setlength{\tabcolsep}{2.7pt}
\setlength{\aboverulesep}{2pt}
\setlength{\belowrulesep}{2pt}
\renewcommand{\arraystretch}{1.12}

\begin{tabular}{l!{\vrule width 0.6pt}ccccc}
\toprule[1.05pt]
Method
& MATH-500 & AIME24 & AIME25 & AMC23 & Avg. \\
\midrule[0.8pt]

\rowcolor{gray!10}
\multicolumn{1}{l!{\vrule width 0.6pt}}{\textbf{\textit{Qwen3.5-9B}}}
& 77.00 & 36.67 & 20.00 & 75.00 & 52.17 \\

GRPO
& 80.20 & 40.00 & 30.00 & 80.00 & 57.55 \\

\rowcolor{cyan!7}
GRPO + \reft
& \textbf{82.80} & \textbf{46.67} & \textbf{33.33}
& \textbf{90.00} & \textbf{63.20} \\

\specialrule{0.8pt}{1pt}{1pt}

\rowcolor{gray!10}
\multicolumn{1}{l!{\vrule width 0.6pt}}{\textbf{\textit{Qwen2.5-14B-Instruct}}}
& 67.80 & 10.00 & 6.67 & 57.50 & 35.49 \\

GRPO
& 75.40 & 13.33 & 10.00 & 65.00 & 40.93 \\

\rowcolor{cyan!7}
GRPO + \reft
& \textbf{76.80} & \textbf{16.67} & \textbf{16.67}
& \textbf{70.00} & \textbf{45.04} \\

\bottomrule[1.1pt]
\end{tabular}
\label{tab:newmodels}
\vspace{-7pt}
\end{wraptable}

\paragraph{Across scale and generation.}
Table~\ref{tab:newmodels} extends the comparison upward to Qwen3.5-9B, a newer-generation model, and to the larger Qwen2.5-14B-Instruct. 
Both gain over four Math~Avg.\ points, +5.65 and +4.11, and improve every benchmark. 
At the other end of the scale, even at 0.5B, where reasoning capacity is limited, \reft{} also raises all three budgets on GSM8K (Appendix~\ref{app:half}). 
The benefit extends across both model scale and generation.

\begin{wraptable}{r}{0.47\linewidth}
\vspace{-15pt}
\centering
\caption{
\textbf{Coverage at a larger sampling budget.}
Pass@64 on GSM8K and Math~Avg.
}
\vspace{-6pt}
\scriptsize
\begingroup
\setlength{\tabcolsep}{10pt}
\setlength{\aboverulesep}{1pt}
\setlength{\belowrulesep}{1pt}
\renewcommand{\arraystretch}{0.92}
\resizebox{\linewidth}{!}{
\begin{tabular}{l!{\vrule width 0.65pt}c!{\vrule width 0.7pt}c}
\toprule[1.1pt]
Method & GSM8K & Math~Avg. \\
\midrule[0.8pt]

\rowcolor{gray!10}
\multicolumn{1}{c!{\vrule width 0.6pt}}{\textbf{\textit{Qwen2.5-3B-Instruct}}}
& 98.18 & 61.95 \\
GRPO
& 98.64 & 64.34 \\
\rowcolor{cyan!7}
GRPO + REFT
& \textbf{99.39} & \textbf{67.82} \\
DAPO
& 97.65 & 65.13 \\
\rowcolor{cyan!7}
DAPO + REFT
& \textbf{98.18} & \textbf{69.18} \\
\specialrule{0.8pt}{2pt}{2pt}

\rowcolor{gray!10}
\multicolumn{1}{c!{\vrule width 0.6pt}}{\textbf{\textit{Llama3.2-3B-Instruct}}}
& 97.42 & 51.61 \\
GRPO
& 97.95 & 55.03 \\
\rowcolor{cyan!7}
GRPO + REFT
& \textbf{98.56} & \textbf{58.99} \\
DAPO
& 97.35 & 56.28 \\
\rowcolor{cyan!7}
DAPO + REFT
& \textbf{98.48} & \textbf{59.34} \\
\specialrule{0.8pt}{2pt}{2pt}

\rowcolor{gray!10}
\multicolumn{1}{c!{\vrule width 0.6pt}}{\textbf{\textit{Qwen2.5-7B-Instruct}}}
& 97.95 & 66.26 \\
GRPO
& 98.33 & 68.95 \\
\rowcolor{cyan!7}
GRPO + REFT
& \textbf{98.94} & \textbf{71.54} \\
DAPO
& \textbf{98.48} & 69.74 \\
\rowcolor{cyan!7}
DAPO + REFT
& 98.41 & \textbf{71.39} \\
\bottomrule[1.1pt]
\end{tabular}
}

\endgroup
\label{tab:pass64}
\vspace{-20pt}
\end{wraptable}

\paragraph{No trade-off against larger-budget discovery.}
RLVR can improve Pass@1 by narrowing the policy. 
As probability concentrates on the trajectories that already succeed, what a larger sampling budget could still discover shrinks with it, and Pass@64 falls even as Pass@1 rises \citep{yue2025does}.
Table~\ref{tab:pass64} shows that \reft{} does not exhibit this trade-off overall.
On GSM8K, Pass@64 is already near ceiling and differences are necessarily small, yet \reft{} still adds $+0.59$ points averaged over the six model-objective pairs.
The effect is clearer on Math~Avg., where the average gain reaches $+3.13$ points.
\Reft{} takes the gain without narrowing what a larger budget can still find.
Full per-benchmark results are in Appendix~\ref{app:full_pass64_results}.

\subsection{Comparison with Alternative Diversification Strategies}
\label{sec:headtohead}
\begin{wraptable}{r}{0.47\linewidth}
\vspace{-13pt}
\centering
\caption{
\textbf{Prior diversification methods and targeted controls.}
Qwen2.5-3B-Instruct on GSM8K with GRPO ($G{=}8$), varying only rollout construction.
Controls raise first-token temperature or apply the same \reft{} sampling scheme to the second token or the highest-entropy position among the first 64 tokens.
}
\vspace{-6pt}
\scriptsize
\setlength{\tabcolsep}{3.5pt}
\setlength{\aboverulesep}{2pt}
\setlength{\belowrulesep}{2pt}
\renewcommand{\arraystretch}{0.93}

\resizebox{\linewidth}{!}{%
\begin{tabular}{l!{\vrule width 0.6pt}ccc}
\toprule[1.1pt]
Method & Pass@1 & Pass@8 & Pass@64 \\
\midrule[0.8pt]

\rowcolor{gray!10}
\multicolumn{1}{l!{\vrule width 0.6pt}}
{\textbf{\textit{Qwen2.5-3B-Instruct}}}
& 80.14 & 94.16 & 98.18  \\

GRPO
& 84.53 & 95.38 & 98.64 \\

\rowcolor{cyan!7}
GRPO + \reft
& \textbf{86.28} & \textbf{96.13} & \textbf{99.39} \\

\specialrule{0.8pt}{2pt}{2pt}

Rollout-MT 
& 85.06 & 95.45 & 98.33 \\

Token-MT 
& 85.67 & 95.68 & 97.95 \\

LATR 
& 84.61 & 96.06 & 97.73 \\

\specialrule{0.8pt}{2pt}{2pt}

First-token $T{=}2.0$
& 85.59 & 95.83 & 98.56 \\

Second-token control
& 85.29 & 95.60 & 98.48 \\

Max-entropy control
& 85.37 & 95.45 & 97.95 \\

\bottomrule[1.1pt]
\end{tabular}%
}
\label{tab:baselines}
\vspace{-12pt}
\end{wraptable}
We compare \reft{} with six alternatives under a fully matched setup (Qwen2.5-3B-Instruct, GSM8K, GRPO, $G{=}8$, LoRA \citep{hu2022lora}), varying only rollout construction.
Rollout-MT and Token-MT implement the multi-temperature strategies of \citet{zhuang2025exploring}, respectively cycling temperatures $\{1.0, \ldots, 1.4\}$ across the rollouts of each group and applying $T{=}1.2$ to the top 40\% high-entropy tokens within a 100K-token rolling window. 
LATR \citep{xing2025lookahead} branches at high-uncertainty positions, and prunes insufficiently divergent branches by edit distance.
Three targeted controls then localize the advantage. 
First-token $T{=}2.0$ raises temperature only at the first token without guaranteeing coverage, and the two positional controls move the same \reft{} sampling scheme to the second token or to the highest-entropy position among the first 64 tokens.

Two patterns stand out in Table~\ref{tab:baselines}. 
(i) Only \reft{} improves at every budget.
Each alternative improves Pass@1 and Pass@8 at the matched budget but lands below the GRPO baseline at Pass@64, reproducing the trade described by \citet{yue2025does}. 
The best alternative changes with the budget, Token-MT at Pass@1, LATR at Pass@8, and First-token $T{=}2.0$ at Pass@64, whereas \reft{} leads at all three.
All these come at the lowest rollout time per step, roughly $1.00\times$ GRPO in our runs versus $1.34$--$4.81\times$ for the prior methods (Appendix~\ref{app:vllm}).
(ii) The controls localize the gain to explicit coverage at the first token.
First-token $T{=}2.0$ falls short despite intervening at the same position, since temperature does not ensure coverage of low-probability alternatives in a small group.
Applying the same \reft~sampling scheme at the second token confirms that early routing itself is beneficial, but it remains clearly below REFT, as the second token is already conditioned on the sampled first token.
Moving further, the max-entropy control performs comparably to or below even the second-token control, as the intervention now lands inside an ongoing reasoning step. 
The effect grows as the intervention approaches the very first choice, a position prior methods have largely overlooked. 
Descriptions and implementation details of the three prior methods are in Appendix~\ref{app:baseline_details}.

\subsection{Code Generation}
\label{sec:code}
\begin{wraptable}{r}{0.57\linewidth}
\vspace{-11pt}
\centering
\caption{
\textbf{Performance on code-generation RLVR.}
GRPO versus GRPO + \reft{} ($G{=}8$) with sandbox-test rewards.
\reft{} improves all twelve comparisons.
}
\vspace{-4pt}
\setlength{\tabcolsep}{3.5pt}
\setlength{\aboverulesep}{2pt}
\setlength{\belowrulesep}{2pt}
\renewcommand{\arraystretch}{1.17}

\resizebox{\linewidth}{!}{%
\begin{tabular}{
l!{\vrule width 0.8pt}
cc!{\vrule width 0.8pt}
cc!{\vrule width 0.8pt}
cc
}
\toprule[1.5pt]
& \multicolumn{2}{c!{\vrule width 0.8pt}}{HumanEval+}
& \multicolumn{2}{c!{\vrule width 0.8pt}}{MBPP+}
& \multicolumn{2}{c}{LCB v4} \\
\cmidrule(lr){2-3}
\cmidrule(lr){4-5}
\cmidrule(lr){6-7}

Method
& Pass@1 & Pass@8
& Pass@1 & Pass@8
& Pass@1 & Pass@8 \\
\midrule[0.8pt]

\rowcolor{gray!10}
\multicolumn{1}{l!{\vrule width 0.8pt}}
{\textbf{\textit{DeepSeek-R1-Distill-Qwen-7B}}}
& 70.5 & 84.8 & 60.8 & 76.2 & 21.0 & 30.9 \\

GRPO
& 72.0 & 86.6
& 61.9 & 78.3
& 22.3 & 32.5 \\

\rowcolor{cyan!7}
GRPO + \reft
& \textbf{73.2} & \textbf{87.8}
& \textbf{65.1} & \textbf{79.9}
& \textbf{23.1} & \textbf{33.7} \\

\specialrule{1.1pt}{2pt}{2pt}

\rowcolor{gray!10}
\multicolumn{1}{l!{\vrule width 0.6pt}}
{\textbf{\textit{Qwen2.5-Coder-3B-Instruct}}}
& 78.4 & 89.6 & 62.5 & 78.3 & 10.0 & 17.5 \\

GRPO
& 79.3 & 90.2
& 63.8 & 79.1
& 10.2 & 17.8 \\

\rowcolor{cyan!7}
GRPO + \reft
& \textbf{82.3} & \textbf{92.1}
& \textbf{64.8} & \textbf{80.7}
& \textbf{10.8} & \textbf{18.8} \\

\bottomrule[1.5pt]
\end{tabular}%
}
\label{tab:code}
\vspace{-10pt}
\end{wraptable}

Table~\ref{tab:code} extends \reft{} to code generation, where it improves on HumanEval+, MBPP+, and LCB v4 for both models, yielding gains in all twelve reported comparisons.
The two models expose different kinds of first semantic choices.
DeepSeek-R1-Distill-Qwen-7B produces an explicit reasoning trace, so \reft{} branches on the first semantic token after \thinktok{}, following the model's official usage recommendation.
As in mathematical reasoning, this token is typically a discourse opener.
Qwen2.5-Coder-3B-Instruct instead generates the program directly, so \reft{} branches on its first semantic code token. 
Unlike the former, the first semantic code token can already influence the program's outer organization, such as whether generation begins with an import, a function definition, or an input-handling construct.
At the same time, it is chosen before most program-specific commitments have been made, so it remains low-load relative to the tokens that follow.
As generation proceeds, syntax, variable bindings, control flow, and earlier implementation decisions increasingly constrain what can follow.
\Reft{} introduces diversity before these dependencies accumulate, varying how a program is organized while leaving the policy to develop each continuation coherently on its own.
Rollout examples are shown in Appendix~\ref{app:qualitative_code}.

\section{Analysis: Why First-Token Coverage Matters}
\label{sec:analysis}
We investigate whether \reft{}'s gains reflect productive exploration rather than changes to first-token concentration alone.
We first examine a first-token loss-masking control, then track how first-token concentration evolves during training.
Finally, we ask whether the additional exploration more often reaches correct completions rather than merely producing more varied failures.

\begin{wraptable}{r}{0.55\linewidth}
\vspace{-12pt}
\centering
\caption{
\textbf{First-token loss masking.}
The mask sets the policy-loss weight of the first token to zero.
GRPO with the mask stays close to GRPO, whereas \reft{} with the same mask improves over masked GRPO at every budget~(Qwen2.5-3B-Instruct on GSM8K with GRPO).
}
\label{tab:mask}
\scriptsize
\setlength{\tabcolsep}{3.5pt}
\setlength{\aboverulesep}{2pt}
\setlength{\belowrulesep}{2pt}
\renewcommand{\arraystretch}{1.15}
\resizebox{\linewidth}{!}{%
\begin{tabular}{l!{\vrule width 0.6pt}ccc}
\toprule[1.1pt]
Method & Pass@1 & Pass@8 & Pass@64 \\
\midrule[0.8pt]
GRPO
& 84.53 & 95.38 & 98.64 \\
GRPO + first-token mask
& 84.61 & 95.75 & 98.64 \\
GRPO + \reft{} + first-token mask
& 85.75 & 95.98 & 99.01 \\
\rowcolor{cyan!7}
GRPO + \reft{}
& \textbf{86.28} & \textbf{96.13} & \textbf{99.39} \\
\bottomrule[1.1pt]
\end{tabular}}
\vspace{-9pt}
\end{wraptable}

\paragraph{\Reft{} remains effective under first-token loss masking.}
One possible explanation for \reft{}'s gains is that it reduces direct reinforcement of dominant first tokens.
To test whether removing these updates alone is sufficient, we retain standard GRPO sampling but set the policy-loss weight of the first token to zero.
The sampling rule and the loss for subsequent tokens are unchanged.
As shown in Table~\ref{tab:mask}, this control remains close to GRPO at Pass@1, Pass@8, and Pass@64.
Thus, suppressing direct first-token updates alone is insufficient to obtain the gains of \reft{}.
By contrast, applying the same mask under \reft{} sampling still improves over the masked GRPO control at every budget.
Although updates on later tokens may still influence the first-token prior indirectly, these results imply that \reft{}'s gains require learning from the first-token-conditioned trajectories introduced by its broader sampling.

\begin{wrapfigure}{r}{0.57\linewidth}
\vspace{-13pt}
\centering

\begin{subfigure}[t]{0.49\linewidth}
    \centering
    \includegraphics[width=\linewidth]{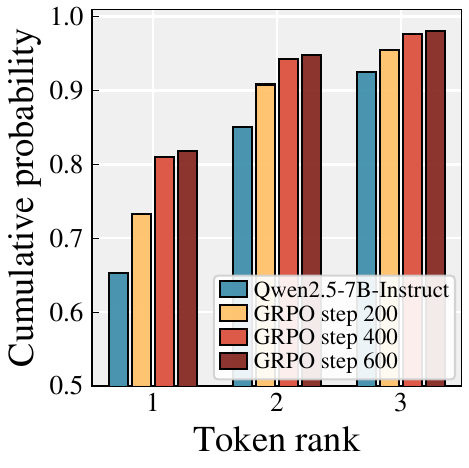}
    \caption{Standard GRPO}
    \label{fig:escalate_grpo}
\end{subfigure}
\hfill
\begin{subfigure}[t]{0.49\linewidth}
    \centering
    \includegraphics[width=\linewidth]{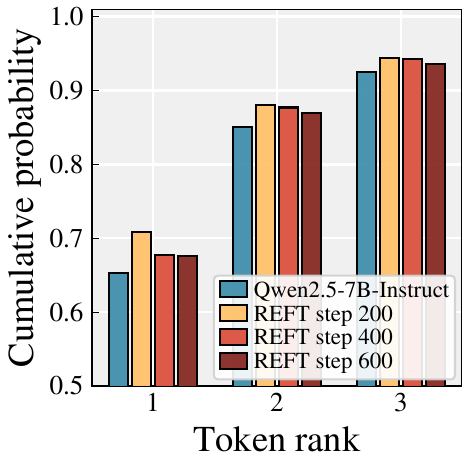}
    \caption{GRPO + REFT}
    \label{fig:escalate_reft}
\end{subfigure}
\vspace{-6pt}
\caption{
\small
\textbf{Less first-token sharpening under REFT.}
Top-$r$ cumulative first-token probability during training (Qwen2.5-7B-Instruct with GSM8K). 
REFT keeps the prior flatter than GRPO.
}
\label{fig:escalate}
\vspace{-20pt}
\end{wrapfigure}

\paragraph{Better performance with less first-token sharpening.}
We next examine the learned policy's first-token distribution.
Under standard GRPO, the top-1 first-token probability increases during training (Figure~\ref{fig:escalate_grpo}), whereas it remains comparatively stable under \reft{} (Figure~\ref{fig:escalate_reft}).
Thus, \reft{} achieves better performance without further sharpening its preference for dominant first tokens.
Together with the masking control, this supports the interpretation that \reft{} benefits from learning on underexplored continuations, rather than merely reinforcing already preferred starting tokens.

\begin{figure*}[!h]
\centering
\begin{subfigure}[t]{0.49\textwidth}
    \centering
    \includegraphics[width=\linewidth]{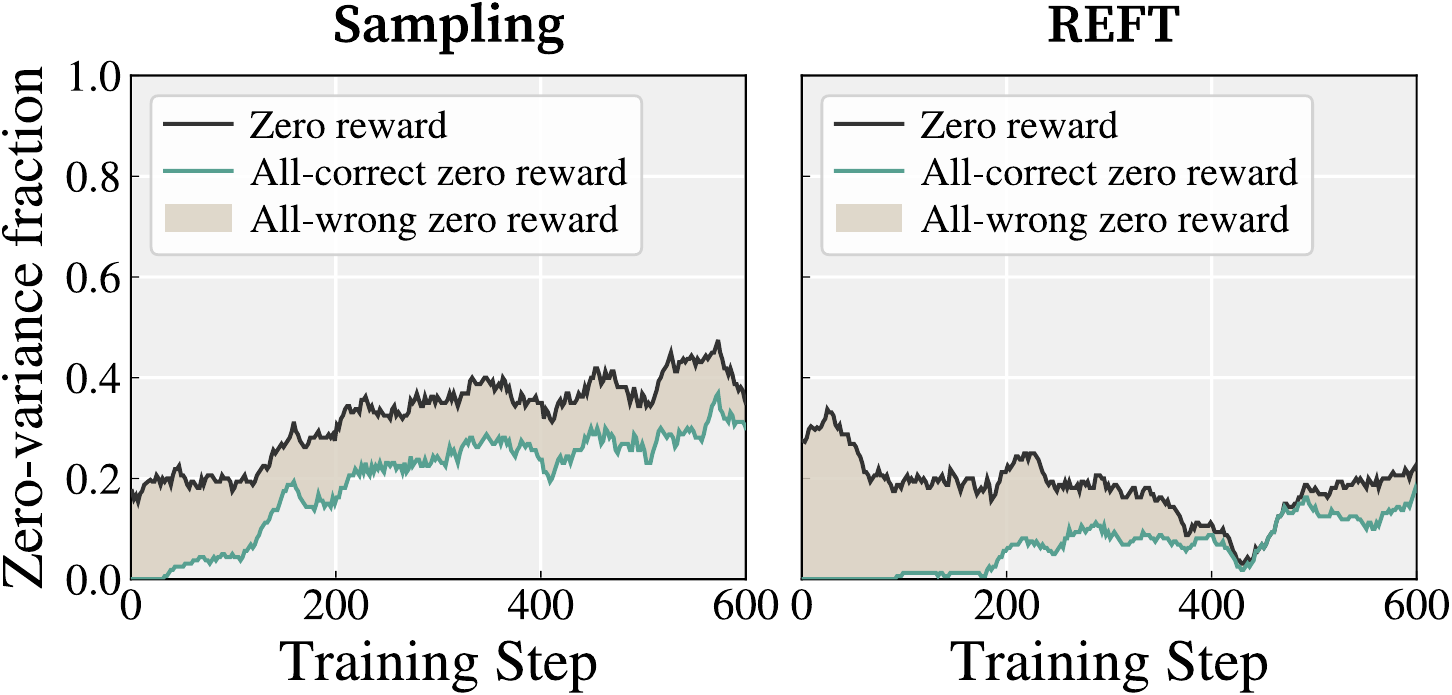}
    \caption{\small Qwen2.5-0.5B-Instruct, GSM8K, DAPO}
    \label{fig:zv_05b}
\end{subfigure}
\hfill
\begin{subfigure}[t]{0.49\textwidth}
    \centering
    \includegraphics[width=\linewidth]{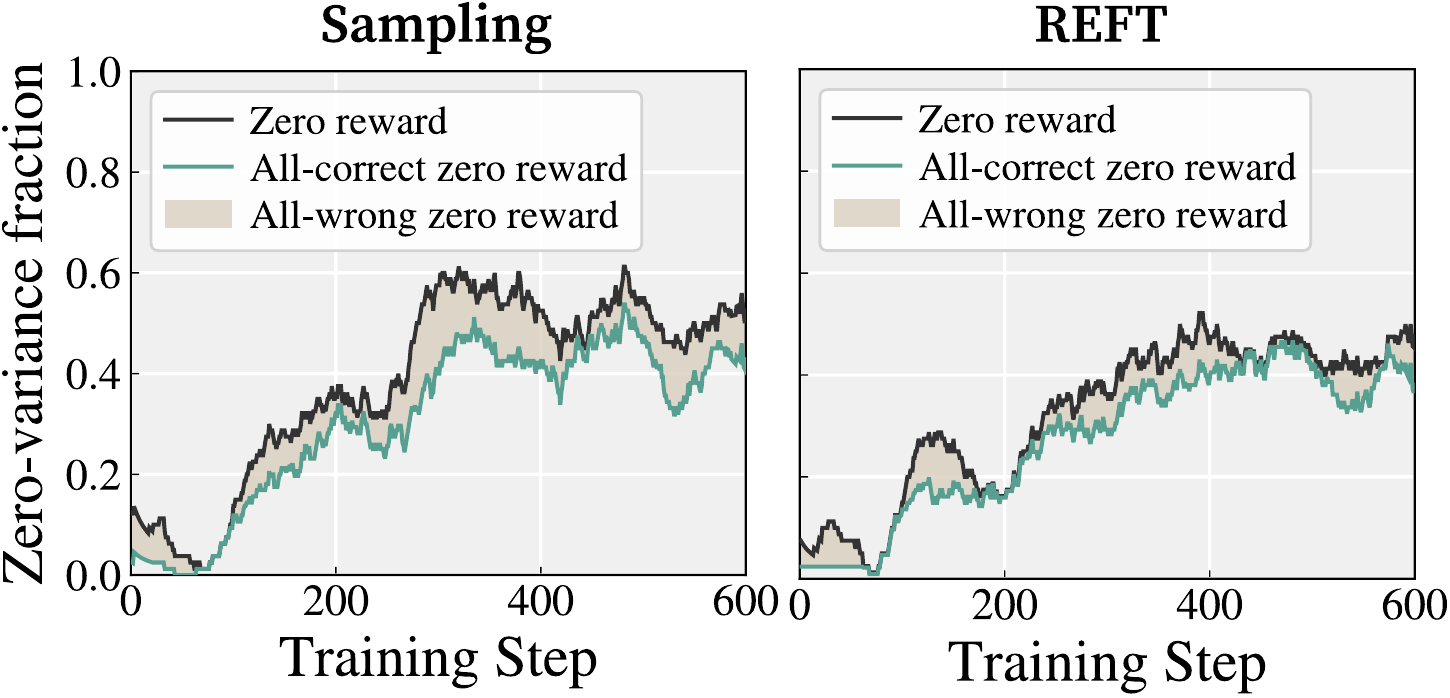}
    \caption{\small Qwen2.5-3B-Instruct, BigMath-Easy, GRPO}
    \label{fig:zv_3b}
\end{subfigure}
\vspace{-4pt}
\caption{\small
\textbf{REFT~reduces all-wrong groups.}
Each panel shows the fraction of zero-variance groups (black) and all-correct subset (green) for standard sampling (left) and REFT (right). The shaded gap is the all-wrong subset.
}
\label{fig:zero_var}
\vspace{-10pt}
\end{figure*}

\paragraph{Broader coverage reaches successful completions.}
Greater diversity need not improve exploration.
A rollout group may contain many different but uniformly unsuccessful trajectories.
We therefore examine how often a group fails to produce any correct completion.
With binary rewards, zero-variance groups are either all-wrong or all-correct.
Although neither provides within-group reward contrast, only an all-wrong group exhausts its rollout budget without a single success.
Figure~\ref{fig:zero_var} shows that \reft{}'s all-wrong rate decreases as training proceeds and falls below that of standard sampling, even when the overall zero-variance rates are comparable.
Thus, \reft{} more often finds at least one correct completion within the same rollout budget, what Pass@$k$ measures, suggesting that its added diversity supports productive exploration rather than merely increasing variation among failure cases.

\vspace{-6pt}


\section{Conclusion}
We study how to diversify a fixed rollout budget in RLVR and identify the first token of the response as a surprisingly efficient exploration site.
Our diagnosis reveals a sharply concentrated first-token prior even though lower-probability alternatives yield distinct continuations with comparable downstream correctness, leaving viable trajectories underexplored.
\Reft{} exploits this mismatch by explicitly covering first tokens drawn uniformly from the policy's top candidates, allocating the fixed rollout budget across them while leaving the rest of the RLVR pipeline unchanged.
Across eight models from 0.5B to 14B in the Qwen, Llama, and DeepSeek-R1-Distill families, \reft{} consistently improves finite-budget performance on mathematical reasoning and code generation, on benchmarks including GSM8K, MATH-500, AIME, HumanEval+, and LiveCodeBench.
The gains hold under both GRPO and DAPO and across Pass@1, Pass@8, and Pass@64, and \reft{} also outperforms alternative diversification strategies under a matched rollout budget.
Overall, useful exploration in RLVR can arise not only at uncertain reasoning pivots, but also from choices the model already treats as settled.

\subsection*{AI use statement}
We used generative AI tool only to assist with writing, such as correcting grammar and polishing wording. All AI-assisted text was reviewed and edited by the authors, who take full responsibility for the final content of this work.

\subsection*{Ethics statement}
This work studies rollout construction for reinforcement learning with verifiable rewards on mathematical reasoning and code generation. It does not involve human subjects, personal data, or crowdsourced annotation. All models and datasets are publicly available and used under their respective licenses (Tables~\ref{tab:model_license} and~\ref{tab:dataset_license}). Generated programs are executed only in a sandboxed environment against provided unit tests. 

\subsection*{Reproducibility statement}
Section~\ref{sec:method} and Algorithm~\ref{alg:reft} specify \reft{} in full, and Appendix~\ref{app:vllm} describes how it is implemented in a standard vLLM pipeline without engine modification. Section~\ref{sec:experimental_setup} summarizes the experimental setup, and Appendix~\ref{app:exp_details} provides the full training and evaluation configurations, including hyperparameters, model and dataset sources, and hardware. Appendix~\ref{app:semantic_diversity} defines the semantic-diversity metric, Appendix~\ref{app:off-policy} details the training objective, and Appendix~\ref{app:seeds} reports three-seed repetitions of the core comparisons. Appendix~\ref{app:baseline_details} specifies the configurations of the compared prior methods, and Appendix~\ref{app:utility} gives the protocol for the general-utility evaluation. We will release our code upon publication.

{
\bibliographystyle{iclr2027_conference}
\bibliography{main}
} 

\medskip


\clearpage
\appendix

\section{Extended First-Token Diagnosis}
\label{app:distributions}
\subsection{Probability and Correctness Across Models} \label{app:extend_distributions}
\begin{figure*}[!h]
    \centering
    \includegraphics[width=\textwidth]{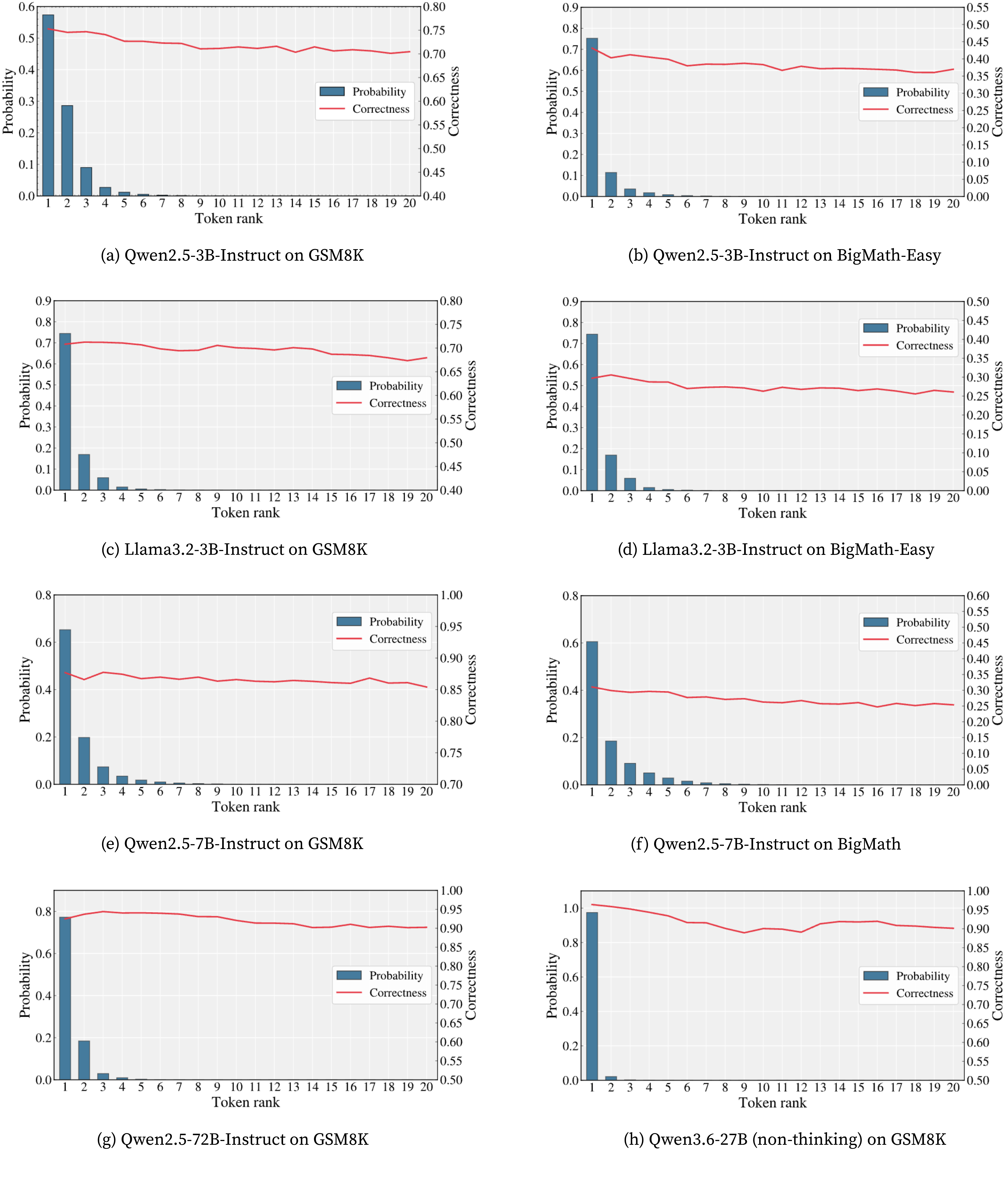}
    \caption{
    \textbf{Probability--correctness decoupling across models and datasets.}
    Bars show mean first-token probability, and lines show forced-first-token rollout correctness.
    Across models and scales, probability drops sharply with token rank, whereas correctness remains comparatively flat.
    }
    \label{fig:extended_distributions}
\end{figure*}

A natural objection is that the pattern in Figure~\ref{fig:main_fig} may be an artifact of a single model--dataset pair.
Figure~\ref{fig:extended_distributions} and Table~\ref{tab:decouplefull} extend the core diagnosis across datasets, model families, scales, and recent checkpoints.
The pattern persists throughout.
Probability mass remains heavily concentrated on the highest-ranked first tokens, while forced-first-token correctness stays comparatively flat across the top-20 alternatives.
This supports the diagnosis in Section~\ref{sec:observations} that the first-token concentration is a recurring phenomenon rather than an artifact of the particular setting in Figure~\ref{fig:main_fig}.

Across the nine model--dataset pairs, the top-1/top-20 probability ratio ranges from about $10^{3}$ to $10^{7}$, while the corresponding correctness gap remains at most 6.3\,pp.
Additionally, the concentration is particularly extreme in several larger and more recent checkpoints (bottom block of Table~\ref{tab:decouplefull}).
Qwen2.5-72B-Instruct exhibits a $1.4\times10^{7}$ top-1/top-20 probability ratio with only a 2.2\,pp correctness gap.
Gemma4-31B-it and Qwen3.6-27B exhibit near-deterministic first-token priors, assigning 99.5\% and 97.4\% of the probability mass to the top-ranked opener, with top-1/top-20 ratios about $3\times10^{6}$ and $2.6\times10^{5}$, respectively.
Even the largest gap in the table, the 6.3\,pp of Qwen3.6-27B, accompanies a five-order-of-magnitude difference in sampling probability.
Extremely sharp first-token priors thus arise in larger and more recent checkpoints without a commensurate separation in correctness.

\begin{table}[h]
\centering
\caption{\textbf{Probability--correctness decoupling across models, datasets, and scales.} $p_1$ and $p_{20}$ are mean probabilities of the rank-1 and rank-20 first tokens, and $C_1 \rightarrow C_{20}$ is forced-first-token rollout correctness at those two ranks. The top-1/top-20 probability ratio ranges from about $10^{3}$ to $10^{7}$, while correctness moves by at most a few points. The bottom block reports larger and more recent models.}
\label{tab:decouplefull}
\small
\resizebox{\linewidth}{!}{%
\begin{tabular}{llccccc}
\toprule
Model & Dataset & $p_1$ & $p_{20}$ & $p_1/p_{20}$ & $C_1 \rightarrow C_{20}$ (\%) & $\Delta C$ (pp) \\
\midrule
Qwen2.5-3B-Instruct & GSM8K & 0.570 & $2.68\times10^{-5}$ & $\approx2.1\times10^{4}$ & $75.29 \rightarrow 70.40$ & $-4.9$ \\
Llama3.2-3B-Instruct & GSM8K & 0.739 & $1.24\times10^{-4}$ & $\approx6.0\times10^{3}$ & $70.83 \rightarrow 67.95$ & $-2.9$ \\
Qwen2.5-7B-Instruct & GSM8K & 0.653 & $1.29\times10^{-5}$ & $\approx5.1\times10^{4}$ & $87.69 \rightarrow 85.41$ & $-2.3$ \\
Qwen2.5-3B-Instruct & BigMath-Easy & 0.711 & $6.38\times10^{-4}$ & $\approx1.1\times10^{3}$ & $43.11 \rightarrow 36.99$ & $-6.1$ \\
Llama3.2-3B-Instruct & BigMath-Easy & 0.753 & $2.88\times10^{-5}$ & $\approx2.6\times10^{4}$ & $28.03 \rightarrow 24.45$ & $-3.6$ \\
Qwen2.5-7B-Instruct & BigMath & 0.606 & $4.62\times10^{-5}$ & $\approx1.3\times10^{4}$ & $31.04 \rightarrow 25.38$ & $-5.7$ \\
\midrule
Qwen2.5-72B-Instruct & GSM8K & 0.773 & $5.49\times10^{-8}$ & $\approx1.4\times10^{7}$ & $92.48 \rightarrow 90.25$ & $-2.2$ \\
Qwen3.6-27B & GSM8K & 0.974 & $3.77\times10^{-6}$ & $\approx2.6\times10^{5}$ & $96.39 \rightarrow 90.13$ & $-6.3$ \\
Gemma4-31B-it & GSM8K & 0.995 & $3.22\times10^{-7}$ & $\approx3.1\times10^{6}$ & $96.66 \rightarrow 96.63$ & $-0.03$ \\
\bottomrule
\end{tabular}%
}
\end{table}

\begin{figure*}[!h]
    \centering
    \includegraphics[width=\linewidth]{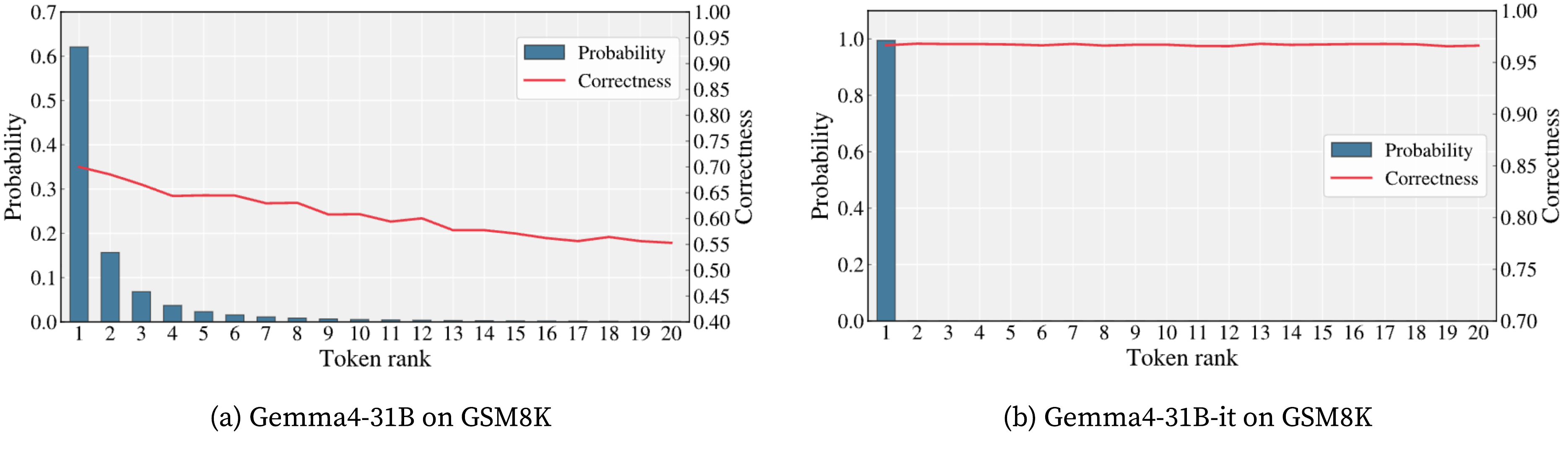}
    \caption{
    \textbf{Gemma4-31B base vs. instruction-tuned on GSM8K.}
    (a) The Gemma4-31B base checkpoint has a relatively soft first-token
    distribution, with correctness decreasing across ranks.
    (b) Gemma4-31B-it concentrates nearly all probability on its top-ranked
    first token, while correctness remains nearly flat across alternative first tokens.
    Bars show first-token probability, and lines show forced-first-token correctness.
    }
    \label{fig:gemma_base_instruct_full}
\end{figure*}

The Gemma4-31B base--instruct comparison provides a particularly clear
within-family contrast (Figure~\ref{fig:gemma_base_instruct_full}).
For the base checkpoint, the first-token distribution is relatively soft,
and forced-first-token correctness declines substantially with rank,
from 70.01\% at rank 1 to 55.30\% at rank 20.
By contrast, after instruction tuning, the distribution becomes extremely
concentrated. 
The instruction-tuned Gemma4-31B-it assigns 0.995 probability to its rank-1 opener,
while the rank-20 probability falls to $3.22\times10^{-7}$.
Despite this dramatic sharpening, correctness is essentially unchanged across the same alternatives, moving from 96.66\% at rank 1 to 96.63\% at rank 20.
Thus, the much sharper prior of the instruction-tuned checkpoint is not
accompanied by a corresponding separation in outcome quality.
Although this comparison does not isolate any particular post-training stage, it shows that a post-trained checkpoint can exhibit dramatically stronger first-token
preference than its base model at the same time as the correctness evidence
for that preference disappears.

\subsection{Per-Prompt First-Token Distributions}
\label{app:per_prompt_first_token}
The aggregate first-token distribution in Section~\ref{sec:observations} averages over many prompts.
This appendix shows that the same concentration phenomenon appears at the individual-prompt level.
For each prompt, we plot the policy's probability over the top-$20$ first-token candidates.
The horizontal axis lists the actual tokens in decreasing-probability order rather than abstract ranks, making the identity of the dominant first tokens visible.
On three GSM8K prompts under Llama3.2-3B-Instruct, almost all probability mass falls on the top few first tokens, with the remaining top-$20$ tokens carrying negligible probability.


\begin{figure}[h]
\centering
\begin{promptbox}[title=GSM8K prompt 0 \quad (gold answer: $18$)]
Janet's ducks lay 16 eggs per day. She eats three for breakfast every morning and bakes muffins for her friends every day with four. She sells the remainder at the farmers' market daily for \$2 per fresh duck egg. How much in dollars does she make every day at the farmers' market?
\end{promptbox}
\vspace{2pt}
\includegraphics[width=0.6\linewidth]{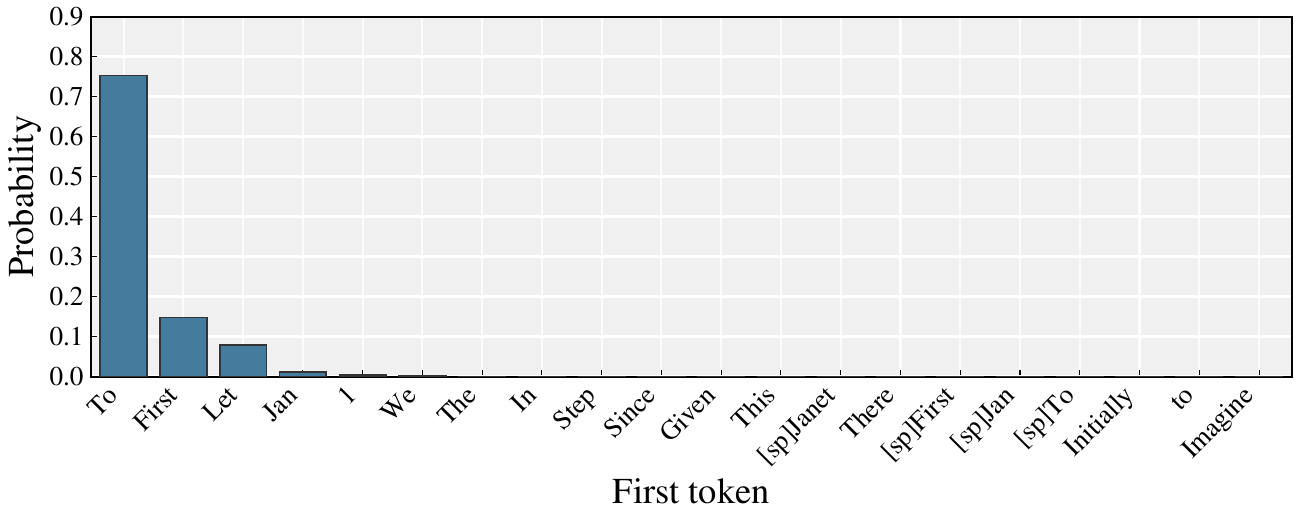}
\vspace{-6pt}
\caption{
\small
\textbf{Top-20 first-token probabilities for GSM8K prompt 0}, under Llama3.2-3B-Instruct.
}
\label{fig:per_prompt_llama3b_gsm8k_0}
\end{figure}

\begin{figure}[h]
\centering
\begin{promptbox}[title=GSM8K prompt 1 \quad (gold answer: $3$)]
A robe takes 2 bolts of blue fiber and half that much white fiber. How many bolts in total does it take?
\end{promptbox}
\vspace{2pt}
\includegraphics[width=0.6\linewidth]{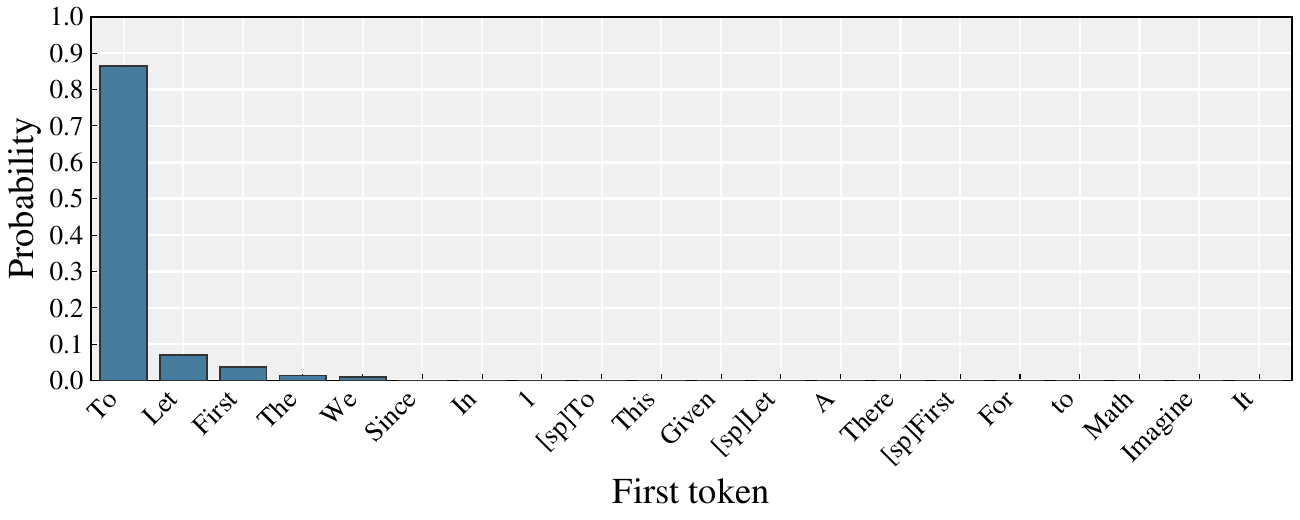}
\vspace{-6pt}
\caption{
\small
\textbf{Top-20 first-token probabilities for GSM8K prompt 1}, under Llama3.2-3B-Instruct.
}
\label{fig:per_prompt_llama3b_gsm8k_1}
\end{figure}

\begin{figure}[h]
\centering
\begin{promptbox}[title=GSM8K prompt 2 \quad (gold answer: $70{,}000$)]
Josh decides to try flipping a house. He buys a house for \$80{,}000 and then puts in \$50{,}000 in repairs. This increased the value of the house by $150\%$. How much profit did he make?
\end{promptbox}
\vspace{2pt}
\includegraphics[width=0.6\linewidth]{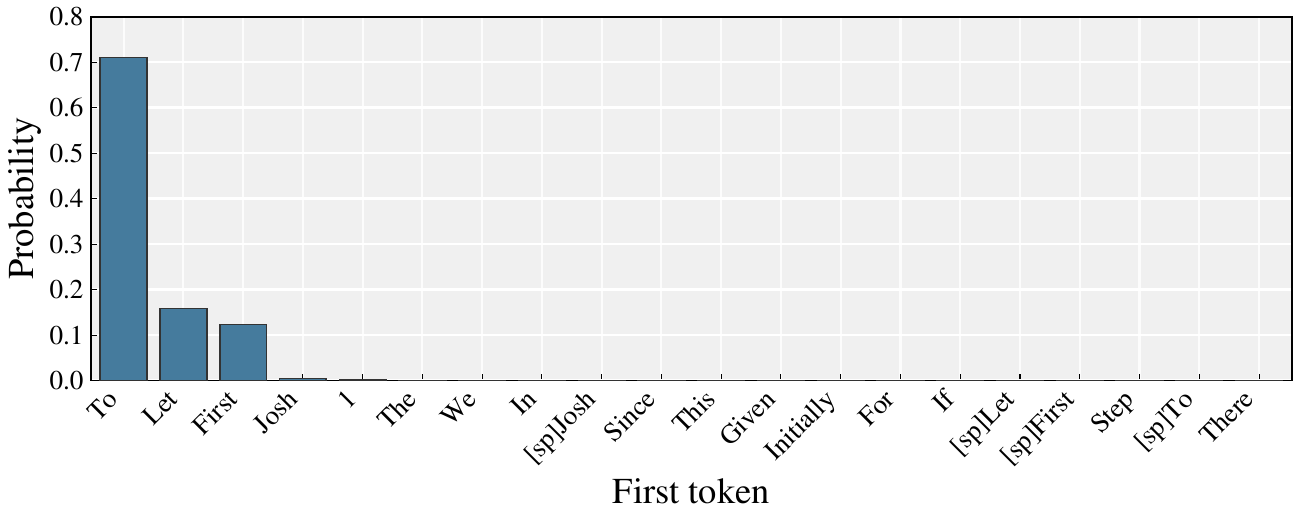}
\vspace{-6pt}
\caption{
\small
\textbf{Top-20 first-token probabilities for GSM8K prompt 2}, under Llama3.2-3B-Instruct.
}
\label{fig:per_prompt_llama3b_gsm8k_2}
\end{figure}

\newpage

\section{Additional Results and Ablations}
\label{app:additional_results}

\subsection{Larger-Budget Coverage}
\label{app:full_pass64_results}
Table~\ref{tab:full_pass64} reports the full per-benchmark Pass@64 results behind the compact summary in Table~\ref{tab:pass64}.
Pass@64 measures larger-budget coverage, namely whether at least one of 64 sampled completions is correct.
Math Avg. averages MATH-500, AIME24, AIME25, and AMC23, excluding GSM8K.
Table~\ref{tab:full_pass64} shows that \Reft{} generally preserves or improves the ability to recover a correct trajectory under a larger sampling budget, rather than trading larger-budget coverage for single-sample accuracy.
\begin{table*}[!h]
\vspace{-10pt}
\centering
\caption{
Pass@64 reasoning performance across models and RLVR methods.
Pass@64 denotes whether at least one of 64 sampled completions is correct.
Math~Avg. excludes GSM8K, and \textbf{bold} marks the higher score.
Each block compares GRPO and DAPO with their \reft{} variants.
}
\setlength{\tabcolsep}{5.5pt}
\setlength{\aboverulesep}{2pt}
\setlength{\belowrulesep}{2pt}
\renewcommand{\arraystretch}{0.82}
\resizebox{0.82\textwidth}{!}{
\begin{tabular}{l!{\vrule width 0.6pt}c!{\vrule width 0.6pt}cccc!{\vrule width 0.6pt}c}
\toprule[1.1pt]
& \multicolumn{1}{c!{\vrule width 0.6pt}}{GSM8K}
& \multicolumn{1}{c}{MATH-500}
& \multicolumn{1}{c}{AIME24}
& \multicolumn{1}{c}{AIME25}
& \multicolumn{1}{c!{\vrule width 0.6pt}}{AMC23}
& \multicolumn{1}{c}{Math~Avg.} \\
\cmidrule(lr){2-2}
\cmidrule(lr){3-3}
\cmidrule(lr){4-4}
\cmidrule(lr){5-5}
\cmidrule(lr){6-6}
\cmidrule(lr){7-7}
Method
& Pass@64
& Pass@64
& Pass@64
& Pass@64
& Pass@64
& Pass@64 \\
\midrule[0.8pt]

\rowcolor{gray!10}
\multicolumn{1}{c!{\vrule width 0.6pt}}{\textbf{\textit{Qwen2.5-3B-Instruct}}}
& 98.18 & 87.80 & 33.33 & 36.67 & 90.00 & 61.95 \\
GRPO
& 98.64 & 88.20 & 36.67 & 40.00 & 92.50 & 64.34 \\
\rowcolor{cyan!7}
GRPO + REFT
& \textbf{99.39} & \textbf{89.60} & \textbf{40.00} & \textbf{46.67} & \textbf{95.00} & \textbf{67.82} \\
DAPO
& 97.65 & 88.00 & 36.67 & 43.33 & 92.50 & 65.13 \\
\rowcolor{cyan!7}
DAPO + REFT
& \textbf{98.18} & \textbf{89.20} & \textbf{43.33} & \textbf{46.67} & \textbf{97.50} & \textbf{69.18} \\
\specialrule{0.8pt}{2pt}{2pt}

\rowcolor{gray!10}
\multicolumn{1}{c!{\vrule width 0.6pt}}{\textbf{\textit{Llama3.2-3B-Instruct}}}
& 97.42 & 80.60 & 30.00 & 13.33 & 82.50 & 51.61 \\
GRPO
& 97.95 & 81.80 & 33.33 & 20.00 & 85.00 & 55.03 \\
\rowcolor{cyan!7}
GRPO + REFT
& \textbf{98.56} & 81.80 & \textbf{36.67} & \textbf{30.00} & \textbf{87.50} & \textbf{58.99} \\
DAPO
& 97.35 & 82.60 & 36.67 & 23.33 & 82.50 & 56.28 \\
\rowcolor{cyan!7}
DAPO + REFT
& \textbf{98.48} & \textbf{83.20} & 36.67 & \textbf{30.00} & \textbf{87.50} & \textbf{59.34} \\
\specialrule{0.8pt}{2pt}{2pt}

\rowcolor{gray!10}
\multicolumn{1}{c!{\vrule width 0.6pt}}{\textbf{\textit{Qwen2.5-7B-Instruct}}}
& 97.95 & 89.20 & 40.00 & 43.33 & 92.50 & 66.26 \\
GRPO
& 98.33 & 90.80 & 43.33 & 46.67 & 95.00 & 68.95 \\
\rowcolor{cyan!7}
GRPO + REFT
& \textbf{98.94} & \textbf{92.00} & \textbf{50.00} & 46.67 & \textbf{97.50} & \textbf{71.54} \\
DAPO
& \textbf{98.48} & 90.60 & 46.67 & 46.67 & 95.00 & 69.74 \\
\rowcolor{cyan!7}
DAPO + REFT
& 98.41 & \textbf{91.40} & 46.67 & \textbf{50.00} & \textbf{97.50} & \textbf{71.39} \\
\bottomrule[1.1pt]
\end{tabular}
}
\label{tab:full_pass64}
\vspace{-10pt}
\end{table*}

\subsection{Results at the Smallest Trained Scale}
\label{app:half}

\begin{wraptable}{r}{0.42\textwidth}
\vspace{-40pt}
\centering
\caption{
GSM8K performance of Qwen2.5-0.5B-Instruct.
}
\vspace{-4pt}
\setlength{\tabcolsep}{6pt}
\setlength{\aboverulesep}{2pt}
\setlength{\belowrulesep}{2pt}
\renewcommand{\arraystretch}{1.15}
\resizebox{\linewidth}{!}{
\begin{tabular}{l!{\vrule width 0.6pt}ccc}
\toprule[1.0pt]
& \multicolumn{3}{c}{GSM8K} \\
\cmidrule(lr){2-4}
Method
& Pass@1 & Pass@8 & Pass@64 \\
\midrule[0.8pt]
\rowcolor{gray!10}
\multicolumn{1}{c!{\vrule width 0.6pt}}{\textbf{\textit{Qwen2.5-0.5B-Instruct}}}
& 43.37 & 77.94 & 90.67 \\
GRPO
& 51.71 & 78.17 & 91.05 \\
\rowcolor{cyan!7}
GRPO + REFT
& \textbf{54.21} & \textbf{79.68} & \textbf{93.71} \\
DAPO
& 52.14 & 78.92 & 92.80 \\
\rowcolor{cyan!7}
DAPO + REFT
& \textbf{53.62} & \textbf{79.45} & \textbf{94.39} \\
\bottomrule[1.0pt]
\end{tabular}
}


\label{tab:qwen05_gsm8k_pass}
\vspace{-27pt}
\end{wraptable}

Table~\ref{tab:qwen05_gsm8k_pass} reports GSM8K results for Qwen2.5-0.5B-Instruct, the smallest model we train. 
Even at 0.5B, where reasoning capacity is limited, \reft{} improves both GRPO and DAPO at Pass@1, Pass@8, and Pass@64. 
The direction of the main results thus holds unchanged at the bottom of the scale range.

\subsection{Candidate Width and Within-Token Replication}
\label{app:ablation_stratify}

\begin{wraptable}{r}{0.30\linewidth}
\vspace{-12pt}
\centering
\begin{minipage}{\linewidth}
\centering
\caption{
Ablation on Top-$N$ with $K=4$.
}
\vspace{-4pt}
\begingroup
\setlength{\tabcolsep}{8pt}
\setlength{\aboverulesep}{1pt}
\setlength{\belowrulesep}{1pt}
\renewcommand{\arraystretch}{0.92}
\resizebox{\linewidth}{!}{
\begin{tabular}{c!{\vrule width 0.6pt}cc}
\toprule[1.0pt]
& \multicolumn{2}{c}{GSM8K} \\
\cmidrule(lr){2-3}
$N$
& Pass@1 & Pass@8 \\
\midrule[0.7pt]
10
& 86.20 & 95.83 \\
\rowcolor{gray!10}
20
& \textbf{86.28} & \textbf{96.13} \\
50
& 85.82 & 95.30 \\
100
& 85.14 & 94.92 \\
\bottomrule[1.0pt]
\end{tabular}
}
\endgroup

\label{tab:ablation_topn}
\end{minipage}

\vspace{8pt}

\begin{minipage}{\linewidth}
\centering
\caption{
Ablation on $K$ with Top-$N=20$.
}
\vspace{-4pt}
\begingroup
\setlength{\tabcolsep}{8pt}
\setlength{\aboverulesep}{1pt}
\setlength{\belowrulesep}{1pt}
\renewcommand{\arraystretch}{0.92}
\resizebox{\linewidth}{!}{
\begin{tabular}{c!{\vrule width 0.6pt}cc}
\toprule[1.0pt]
& \multicolumn{2}{c}{GSM8K} \\
\cmidrule(lr){2-3}
$K$ & Pass@1 & Pass@8 \\
\midrule[0.7pt]
1 & 84.23 & 95.15 \\
2 & 85.51 & 95.68 \\
\rowcolor{gray!10}
4 & \textbf{86.28} & \textbf{96.13} \\
8 & 85.82 & 95.38 \\
\bottomrule[1.0pt]
\end{tabular}
}
\endgroup
\label{tab:ablation_k}
\end{minipage}
\end{wraptable}

We further verify that \Reft{} is robust to variations in the candidate width Top-$N$ and the number of selected first tokens $K$.
Tables~\ref{tab:ablation_topn} and~\ref{tab:ablation_k} report ablations on GSM8K with Qwen2.5-3B-Instruct, where the GRPO baseline achieves $84.53$ Pass@1 and $95.38$ Pass@8.
Most \Reft{} variants improve over the baseline, confirming robustness to these hyperparameter choices.
Expanding the pool to Top-$50$ or Top-$100$ remains competitive in Pass@1 but gradually reduces Pass@8, suggesting that an overly wide candidate pool can spend the finite rollout budget on too many exceedingly rare first tokens.
Fixing Top-$N=20$, increasing $K$ from $1$ to $4$ raises Pass@1 and Pass@8, showing the benefit of covering multiple first token regions within the same group.
Increasing $K$ further to $8$ reduces the gain on Pass@8, although Pass@1 remains above the baseline.
With $G=8$, this allocates only one continuation per selected first token, so a viable first token can be misjudged from a single unlucky trace.
Setting $K<G$ is therefore not necessary, but it deepens the sample behind each selected token.
Our default $K=4$ assigns two continuations per first token, balancing broader first-token coverage with enough replication to stabilize the rollout signal. 

\subsection{Three-Seed Repetitions}
\label{app:seeds}

REFT improves the mean in every repeated comparison across the three models, raising Math~Avg. by $4.24$, $3.10$, and $5.07$ points. 
The ranges are well separated on GSM8K, AMC23, and Math~Avg.
On the 30-question AIME sets, where a single problem shifts the score by $3.33$ points, the ranges partly overlap on the 3B models, but the mean direction remains positive.
On Qwen2.5-7B-Instruct both sets separate, with means rising by $6.67$ and $4.45$ points.
Thus, the gains are consistent across seeds, not driven by one favorable run. 

\begin{table}[h]
\vspace{-2pt}
\centering
\caption{Three-seed repetitions of the core comparisons. Pass@1 as mean$\pm$SD over three independent training seeds, under the GRPO objective.}
\label{tab:seeds}
\newcommand{\msd}[2]{$#1_{\pm#2}$}
\scriptsize
\setlength{\tabcolsep}{4.5pt}
\setlength{\aboverulesep}{2pt}
\setlength{\belowrulesep}{2pt}
\renewcommand{\arraystretch}{1.0}
\resizebox{0.95\textwidth}{!}{%
\begin{tabular}{c!{\vrule width 0.6pt}c!{\vrule width 0.6pt}c!{\vrule width 0.6pt}cccc!{\vrule width 0.6pt}c}
\toprule[1.0pt]
Model & Method & GSM8K & MATH-500 & AIME24 & AIME25 & AMC23 & Math Avg. \\
\midrule[0.8pt]
\multirow{2}{*}{Qwen2.5-3B-Instruct} & GRPO & \msd{84.56}{0.27} & \msd{59.73}{0.50} & \msd{8.89}{1.92} & \msd{4.44}{1.92} & \msd{43.33}{1.44} & \msd{29.10}{0.25} \\
& GRPO + \reft & \msd{86.23}{0.31} & \msd{61.13}{0.42} & \msd{15.56}{1.92} & \msd{6.67}{3.33} & \msd{50.00}{2.50} & \msd{33.34}{1.48} \\
\midrule
\multirow{2}{*}{Llama3.2-3B-Instruct} & GRPO & \msd{75.74}{0.35} & \msd{35.53}{0.50} & \msd{5.56}{1.92} & \msd{2.22}{1.92} & \msd{21.67}{1.44} & \msd{16.24}{0.81} \\
& GRPO + \reft & \msd{77.69}{0.31} & \msd{36.80}{0.40} & \msd{6.67}{3.33} & \msd{5.56}{1.92} & \msd{28.33}{2.89} & \msd{19.34}{1.52} \\
\midrule
\multirow{2}{*}{Qwen2.5-7B-Instruct} & GRPO & \msd{90.95}{0.32} & \msd{72.27}{0.83} & \msd{14.44}{1.92} & \msd{12.22}{1.92} & \msd{54.17}{3.82} & \msd{38.27}{1.67} \\
& GRPO + \reft & \msd{91.84}{0.31} & \msd{73.07}{0.50} & \msd{21.11}{1.92} & \msd{16.67}{0.00} & \msd{62.50}{2.50} & \msd{43.34}{0.80} \\
\bottomrule[1.0pt]
\end{tabular}}
\vspace{-10pt}
\end{table}

\subsection{General Utility Retention}
\label{app:utility}

Since \reft{} changes how training rollouts begin, we additionally check whether the trained policy retains its general capabilities outside the RLVR training domain, such as instruction following and general knowledge.
We evaluate the final checkpoints on three general-purpose benchmarks using standard inference, with no \reft{} intervention at evaluation time.

IFEval~\citep{zhou2023instruction} measures instruction following with prompt-level strict accuracy under zero-shot greedy decoding, and being generation-based, it is the most sensitive of the three to how a response begins.
BBH~\citep{suzgun2023challenging} uses 3-shot normalized accuracy over 24 subtasks, and MMLU-Pro~\citep{wang2024mmlu} uses 5-shot multiple-choice accuracy.
Table~\ref{tab:utility} compares the non-trained baseline model, GRPO, and \reft{} for Qwen2.5-3B-Instruct trained on GSM8K and BigMath-Easy and for Qwen2.5-14B-Instruct trained on BigMath.
 
\begin{table*}[!h]
\vspace{-10pt}
\centering
\caption{\textbf{General utility after RLVR training.}
All checkpoints are evaluated with standard inference and no \reft{} intervention.
Gray rows report the original model before RLVR training, and parentheses give the RLVR training dataset.
Bold marks the higher score between GRPO and GRPO+\reft{} within each training setting.}
\label{tab:utility}
\small
\setlength{\tabcolsep}{6pt}
\setlength{\aboverulesep}{2pt}
\setlength{\belowrulesep}{2pt}
\renewcommand{\arraystretch}{1.12}
\begin{tabular}{l!{\vrule width 0.6pt}ccc}
\toprule[1.05pt]
Method & IFEval & BBH & MMLU-Pro \\
& (strict) & (3-shot) & (5-shot) \\
\midrule[0.8pt]
 
\rowcolor{gray!10}
\multicolumn{1}{l!{\vrule width 0.6pt}}{\textbf{\textit{Qwen2.5-3B-Instruct}}}
& 59.89 & 46.61 & 32.71 \\
 
GRPO (GSM8K)
& \textbf{60.81} & 46.59 & \textbf{32.92} \\
\rowcolor{cyan!7}
GRPO + \reft{} (GSM8K)
& \textbf{60.81} & \textbf{46.68} & 32.75 \\
 
GRPO (BigMath-Easy)
& 59.15 & 46.55 & \textbf{33.21} \\
\rowcolor{cyan!7}
GRPO + \reft{} (BigMath-Easy)
& \textbf{60.07} & \textbf{46.75} & 33.05 \\
 
\specialrule{0.8pt}{1pt}{1pt}
 
\rowcolor{gray!10}
\multicolumn{1}{l!{\vrule width 0.6pt}}{\textbf{\textit{Qwen2.5-14B-Instruct}}}
& 78.56 & 63.88 & 49.08 \\
 
GRPO (BigMath)
& 78.37 & 63.70 & 48.99 \\
\rowcolor{cyan!7}
GRPO + \reft{} (BigMath)
& \textbf{79.85} & \textbf{63.74} & \textbf{49.09} \\
 
\bottomrule[1.1pt]
\end{tabular}
\vspace{-5pt}
\end{table*}

Across the three settings, \reft{} shows no noticeable degradation relative to the original model, scoring within $0.2$ points of it or higher on every benchmark.
Compared with GRPO, \reft{} performs comparably on all three benchmarks.
Training on math rollouts thus changes general utility little, and \reft{} improves target-domain performance while retaining it at the level of standard GRPO.

\section{Experimental Details}
\label{app:exp_details}

\subsection{RL Training and Evaluation Details}
\label{app:training_details}

We evaluate \Reft{} as a drop-in rollout-sampling modification on top of DAPO~\citep{yudapo} and GRPO~\citep{shao2024deepseekmath}.  
Table~\ref{tab:model_license} summarizes the models used for RLVR
training and diagnosis, together with the sentence encoder used in our
semantic-diversity analysis.

\renewcommand{\arraystretch}{1.3}
\begin{table}[!htbp]
\centering
\caption{
Summary of models used in this paper.
We report the source and license of the base models, together with the sentence encoder used in our semantic-diversity analysis.
}
\label{tab:model_license}
\setlength{\tabcolsep}{7pt}
\resizebox{0.97\linewidth}{!}{%
\begin{tabular}{l!{\vrule width 0.6pt}l l}
\toprule[1.0pt]
\textbf{Model} & \textbf{Source} & \textbf{License} \\
\midrule[0.8pt]
Qwen2.5-0.5B-Instruct
& Qwen~\citep{yang2024qwen25}
& Apache License 2.0 \\

Qwen2.5-3B-Instruct
& Qwen~\citep{yang2024qwen25}
& Qwen Research License \\

Qwen2.5-7B-Instruct
& Qwen~\citep{yang2024qwen25}
& Apache License 2.0 \\

Qwen2.5-14B-Instruct
& Qwen~\citep{yang2024qwen25}
& Apache License 2.0 \\

Qwen2.5-72B-Instruct
& Qwen~\citep{yang2024qwen25}
& Qwen License \\

Qwen3.5-9B
& Qwen~\citep{qwen35blog}
& Apache License 2.0 \\

Llama3.2-3B-Instruct
& Meta~\citep{grattafiori2024llama}
& Llama 3.2 Community License \\

Qwen3.6-27B
& Qwen~\citep{qwen3.6-27b}
& Apache License 2.0 \\

Gemma4-31B
& Google~\citep{team2026gemma}
& Apache License 2.0 \\

Gemma4-31B-it
& Google~\citep{team2026gemma}
& Apache License 2.0 \\

\midrule[0.8pt]

DeepSeek-R1-Distill-Qwen-7B
& DeepSeek-AI~\citep{guo2025deepseek}
& MIT \\

Qwen2.5-Coder-3B-Instruct
& Qwen~\citep{hui2024qwen2}
& Qwen Research License \\

\midrule[0.8pt]

all-mpnet-base-v2
& Sentence Transformers~\citep{reimers2019sentence,song2020mpnet}
& Apache License 2.0 \\
\bottomrule[1.0pt]
\end{tabular}%
}
\end{table}

\paragraph{\Reft{} configuration.}
Unless otherwise stated, \Reft{} uses candidate width $N=20$ and selected-token count $K=4$.  
For each prompt, four distinct first semantic tokens are sampled uniformly from the policy's top-20 valid candidates, and the fixed rollout budget is allocated evenly across them.
Thus, settings with $G=8$ (GSM8K and code generation) use two continuations per selected first token, and BigMath settings with $G=16$ use four.
The choice $N=20$ follows the diagnostic result in Section~\ref{sec:observations}, where rollout correctness remains nearly flat across the top-20 first-token ranks even though their probabilities differ by orders of magnitude.  
The choice $K=4$ provides coverage of multiple continuation regions while still assigning replicated continuations to each selected first token, reducing sensitivity to a single unlucky rollout.
After the selected first token, the remainder of each rollout is generated with the same continuation decoder as the corresponding baseline.
Ablations and further discussion are given in Appendix~\ref{app:ablation_stratify}.

\paragraph{Mathematical reasoning.}
We use GSM8K~\citep{cobbe2021training}, BigMath-Easy, and BigMath~\citep{albalak2025big} for mathematical reasoning.
GSM8K~\citep{cobbe2021training} contains 7,500 elementary-grade math problems and is used with rollout group size $G=8$ for all four GSM8K-trained models.
For the models with 7 billion parameters or more, we train on the BigMath~\citep{albalak2025big} levels 3--5 subset, containing 121,503 problems.  
For the 3B-scale models, Qwen2.5-3B-Instruct and Llama3.2-3B-Instruct, we construct BigMath-Easy by restricting BigMath to levels 1--3, yielding 127,945 problems.  
This follows prior evidence that removing the hardest problems can provide a more effective training signal for smaller models~\citep{albalak2025big}.  
Both BigMath subsets use rollout group size $G=16$.
Maximum completion length is $4{,}096$ tokens for GSM8K and $8{,}192$ tokens for BigMath and BigMath-Easy.
Unless explicitly varied by a baseline, all math rollout sampling uses temperature $1.0$ and top-$p=1.0$~\citep{huopen,huang2025qerl,gu2026qarl,fang2026allocate}.
The verifier reward combines answer accuracy and format, with weights $(2.0,0.2)$ on GSM8K and $(1.0,0.2)$ on BigMath and BigMath-Easy.
The same reward function and weights are used for the corresponding baseline and \Reft{} runs.
Training runs for up to 1,000 steps using LoRA~\citep{hu2022lora} with rank $32$ and DeepSpeed ZeRO-3~\citep{rajbhandari2020zero} through Hugging Face Accelerate for memory efficiency.
Table~\ref{tab:train_hyperparameters} summarizes the training
configuration.

\paragraph{Code generation.}
We train DeepSeek-R1-Distill-Qwen-7B and Qwen2.5-Coder-3B-Instruct with GRPO ($G=8$) on
\texttt{open-r1/}\allowbreak \texttt{verifiable-coding-problems-python}\allowbreak
\texttt{\_decontaminated-tested-shuffled}~\citep{openr1codeset}, a decontaminated, test-verified collection of Python programming problems.
Generated programs are executed against the provided test cases using the Morph sandbox backend, and the reward is the fraction of tests passed.
DeepSeek-R1-Distill-Qwen-7B generates an explicit reasoning trace, whereas Qwen2.5-Coder-3B-Instruct generates Python directly without an explicit reasoning marker.
Accordingly, \Reft{} branches on the first semantic token after \thinktok{} for DeepSeek and on the first semantic code token for Qwen2.5-Coder.
Within each model, GRPO and GRPO+\Reft{} use identical optimization and continuation-decoding settings.
All code-training rollouts are generated with vLLM~\citep{kwon2023efficient} in colocate mode.
Table~\ref{tab:code_hparams} summarizes the code-training configuration.

\begin{table}[!htbp]
\centering
\caption{
Hyperparameters for math RLVR training.
}
\label{tab:train_hyperparameters}
\setlength{\tabcolsep}{6pt}
\renewcommand{\arraystretch}{1.0}
\begin{tabular}{@{}ll@{}}
\toprule
\textbf{Hyperparameter} & \textbf{Value} \\
\midrule
Optimizer & AdamW-8bit \\
Training batch size & 128 \\
Policy learning rate & $5\mathrm{e}{-6}$ \\
LoRA rank & 32 \\
Rollouts per prompt & 8 for GSM8K, 16 for BigMath / BigMath-Easy \\
Max response length & 4096 for GSM8K, 8192 for BigMath / BigMath-Easy \\
Rollout temperature & 1.0 \\
Rollout top-$p$ & 1.0 \\
Clip range $\epsilon_{\mathrm{low}}, \epsilon_{\mathrm{high}}$ & $0.2$, $0.28$ \\
Policy updates per rollout & 4 for GRPO, 1 for DAPO \\
Reward weights & GSM8K: Accuracy $2.0$, Format $0.2$ \\
               & BigMath / BigMath-Easy: Accuracy $1.0$, Format $0.2$ \\
Training step & 1,000 steps \\
\bottomrule
\end{tabular}
\end{table}

\begin{table}[!htbp]
\centering
\caption{
Hyperparameters for code-generation RLVR training.
}
\label{tab:code_hparams}
\setlength{\tabcolsep}{6pt}
\renewcommand{\arraystretch}{1.0}
\begin{tabular}{@{}ll@{}}
\toprule
\textbf{Hyperparameter} & \textbf{Value} \\
\midrule
RL objective & GRPO \\
Optimizer & AdamW-8bit \\
Policy learning rate & $5\mathrm{e}{-6}$ \\
LoRA rank & 32 \\
Rollouts per prompt & 8 \\
Max prompt length & 2048 \\
Max response length & 8192 \\
Rollout temperature
    & Qwen2.5-Coder: $1.0$, DeepSeek-R1-Distill: $0.6$ \\
Rollout top-$p$
    & Qwen2.5-Coder: $1.0$, DeepSeek-R1-Distill: $0.95$ \\
Clip range $\epsilon_{\mathrm{low}}, \epsilon_{\mathrm{high}}$
    & $0.2$, $0.28$ \\
Policy updates per rollout & 1 \\
Training step & 1,000 steps \\
\bottomrule
\end{tabular}
\end{table}

\paragraph{Evaluation datasets and inference.}
GSM8K-trained models are evaluated on the full GSM8K test set~\citep{cobbe2021training}.
BigMath and BigMath-Easy trained models are evaluated on the full evaluation sets of MATH-500~\citep{lightman2023let}, AIME 2024~\citep{aime24}, AIME 2025~\citep{aime25}, and AMC 2023~\citep{li2024numinamath}.
Math inference uses vLLM~\citep{kwon2023efficient} with temperature $0.6$ and top-$p=0.95$, and we report Pass@1, Pass@8, and Pass@64.
For code generation, we evaluate HumanEval+ and MBPP+ using EvalPlus~\citep{liu2023your} and LiveCodeBench v4 using execution-based correctness~\citep{jain2025livecodebench}.
Code inference uses the same sampling configuration, and we report Pass@1 and Pass@8.
Each Pass@$k$ value is computed from $k$ sampled responses per problem and indicates whether at least one response is correct.
All results are reported at the final checkpoint of each run. 

\begin{table}[!htbp]
\centering
\caption{
Summary of datasets used in this paper.
We report the source and license metadata for both training and evaluation datasets.
}
\label{tab:dataset_license}
\setlength{\tabcolsep}{7pt}
\renewcommand{\arraystretch}{1.05}
\resizebox{0.92\linewidth}{!}{%
\begin{tabular}{l!{\vrule width 0.6pt}l l}
\toprule[1.0pt]
\textbf{Dataset} & \textbf{Source} & \textbf{License} \\
\midrule[0.8pt]
GSM8K
& OpenAI~\citep{cobbe2021training}
& MIT \\

BigMath
& SynthLabsAI Big-Math-RL-Verified~\citep{albalak2025big}
& Apache License 2.0 \\

MATH-500
& OpenAI PRM800K / HuggingFaceH4~\citep{lightman2023let}
& MIT (source repo) \\

AIME 2024
& math-ai~\citep{aime24}
& Apache License 2.0 \\

AIME 2025
& math-ai~\citep{aime25}
& Apache License 2.0 \\

AMC 2023
& NuminaMath eval suite~\citep{li2024numinamath}
& Apache License 2.0 \\

Verifiable Python Problems
& open-r1~\citep{openr1codeset}
& Not specified \\

HumanEval+
& EvalPlus~\citep{liu2023your}
& Apache License 2.0 \\

MBPP+
& EvalPlus~\citep{liu2023your}
& Apache License 2.0 \\

LiveCodeBench v4
& LiveCodeBench~\citep{jain2025livecodebench}
& MIT \\

IFEval
& Google Research~\citep{zhou2023instruction}
& Apache License 2.0 \\
 
BBH
& BIG-Bench Hard~\citep{suzgun2023challenging}
& MIT \\
 
MMLU-Pro
& TIGER-Lab~\citep{wang2024mmlu}
& MIT \\ 

\bottomrule[1.0pt]
\end{tabular}%
}
\end{table}

\paragraph{Hardware.}
We train Qwen2.5-7B/14B-Instruct, Qwen3.5-9B, and DeepSeek-R1-Distill-Qwen-7B on $4\times$ NVIDIA B200 GPUs.
All remaining models are trained on $8\times$ NVIDIA L40S GPUs.
For consistency, all runs for the same base model use the same hardware configuration.

\subsection{Baseline Implementations for the Head-to-Head Comparison}
\label{app:baseline_details}
All alternatives in Section~\ref{sec:headtohead} share the data, prompts, reward, rollout budget ($G{=}8$), GRPO objective, LoRA configuration, and optimization settings of Appendix~\ref{app:training_details}, and differ only in how training rollouts are constructed.
The sampling settings below refer to training rollouts, and all methods are evaluated with the common decoding setting of Appendix~\ref{app:training_details}.
 
\paragraph{Rollout-MT.}
Rollout-level multi-temperature sampling of \citet{zhuang2025exploring} diversifies a group by sampling its rollouts at different temperatures, so conservative and more exploratory rollouts coexist in the same group rather than all sharing a single temperature.
We use the temperature set $\{1.0, 1.1, 1.2, 1.3, 1.4\}$, the in-range mixture that the authors identify as the stable regime.
Following the official implementation, temperatures are assigned cyclically to the $G{=}8$ rollouts of each group, giving $(1.0, 1.1, 1.2, 1.3, 1.4, 1.0, 1.1, 1.2)$, and each rollout keeps its temperature for every token.
 
\paragraph{Token-MT.}
Token-level temperature scheduling of \citet{zhuang2025exploring} concentrates exploration on high-entropy positions, raising the temperature only where the policy is uncertain.
Thus, rollouts are more likely to diverge at these decision points, and low-entropy tokens are sampled as usual.
We use the default configuration of the official implementation, with $T_{\mathrm{low}}{=}1.0$, $T_{\mathrm{high}}{=}1.2$, a high-entropy ratio of $40\%$, and an entropy window over the most recent $100{,}000$ generated token positions.
At every decoding step, the entropy of the full-vocabulary next-token distribution is computed from the untempered logits and added to the window.
Positions at or above the $60$th percentile of the window, roughly the top $40\%$, are sampled at $T_{\mathrm{high}}$, and all others at $T_{\mathrm{low}}$.

\paragraph{LATR.}
Lookahead tree-based rollouts of \citet{xing2025lookahead} enforce diversity structurally rather than through sampling noise.
At uncertain positions, the group branches onto alternative candidate tokens, each new branch is extended for a short lookahead, and branches that remain too similar to their parent are pruned, thus only genuinely divergent trajectories survive in the group.
We use hyperparameters reported in the paper which coincide with the defaults of the official implementation.
At every generation step, each active branch continues with its top-1 token, and a new branch is opened on the top-2 token when its probability both exceeds the absolute threshold $\tau_{\mathrm{abs}}{=}0.25$ and lies within $\tau_{\mathrm{rel}}{=}0.15$ of the top-1 probability.
Each new branch is compared with its parent after $20$, $30$, and $50$ lookahead tokens, and it is pruned together with its descendants if the normalized token edit distance falls below $\tau_{\mathrm{ed}}{=}0.4$ at any of the three windows.
We also adopt the hybrid rollout schedule of the paper, which assigns $k_{\mathrm{LATR}}{=}\lfloor \eta_t G\rceil$ rollouts of each group to the tree and the rest to standard stochastic sampling, with $\eta_t{=}0.995^{\,t}$ at training step $t$ as reported for the math setting.
The tree is therefore capped at $k_{\mathrm{LATR}}$ live branches per prompt, and once the cap is reached, lower-probability branches are suppressed.
Following the official math training script, tree construction stops early once the branch set is full, using a fullness threshold that starts at $0.98$ and decays linearly with rate $0.05$ over the generation, and in any case after $1{,}024$ tokens.
After building the tree prefix, colocated vLLM completes every sequence at $T{=}1.0$ and top-$p{=}1.0$.
The remaining $G-k_{\mathrm{LATR}}$ rollouts are sampled by the same generator up to the same prefix length before the vLLM continuation.

\subsection{Runtime Overhead}
\label{app:runtime_overhead}
\paragraph{Runtime efficiency.}
\Reft{} does not increase the rollout budget.
It only changes how the first token is allocated within the same rollout group.
As shown in Table~\ref{tab:runtime_length}, this makes the method essentially cost-free in practice, and the measured GPU-hours are slightly lower than those of standard sampling.
In the profiled runs, \Reft{} often reaches higher accuracy (Tables~\ref{tab:main_pass1_pass8} and~\ref{tab:qwen05_gsm8k_pass}) with shorter average completions, reducing generation length enough to offset the small overhead of first-token routing.

\begin{table}[!htbp]
\vspace{-6pt}
\centering
\caption{
Runtime and completion-length changes of \Reft{}.
Negative relative differences indicate that \Reft{} is lower than standard sampling. Mean relative differences are computed from the mean values in each column. 
}
\label{tab:runtime_length}
\vspace{-2pt}
\begingroup
\small
\setlength{\tabcolsep}{3.8pt}
\setlength{\aboverulesep}{1.5pt}
\setlength{\belowrulesep}{1.5pt}
\renewcommand{\arraystretch}{1.0}
\begin{tabular}{l!{\vrule width 0.6pt}c!{\vrule width 0.6pt}ccc}
\toprule[1.0pt]
& \multicolumn{1}{c!{\vrule width 0.6pt}}{Runtime}
& \multicolumn{3}{c}{Completion Length} \\
\cmidrule(lr){2-2}
\cmidrule(lr){3-5}
Configuration
& GPU-hour Rel. $\Delta$
& Sampling Avg. & \Reft{} Avg. & Rel. $\Delta$ \\
\midrule[0.8pt]
GSM8K, Qwen2.5-3B-Instruct, DAPO
& $-0.02\%$ & 245.55 & 235.87 & $-3.94\%$ \\
BigMath-Easy, Qwen2.5-3B-Instruct, GRPO
& $-0.80\%$ & 366.25 & 342.33 & $-6.53\%$ \\
BigMath-Easy, Qwen2.5-3B-Instruct, DAPO
& $-0.42\%$ & 353.18 & 333.26 & $-5.64\%$ \\
\midrule[0.8pt]
\textbf{Mean}
& $\mathbf{-0.41\%}$ & \textbf{321.66} & \textbf{303.82} & $\mathbf{-5.55\%}$ \\
\bottomrule[1.0pt]
\end{tabular}
\endgroup
\vspace{-8pt}
\end{table}

\subsection{\reft{} in vLLM-Style Decoding}
\label{app:vllm}
\reft{} requires no engine modification. 
Candidate extraction uses standard log-probability outputs, and forcing a first token amounts to appending one token id to the prompt ids of its assigned rollouts. 
Measured rollout time per step is $\approx$1.00$\times$ standard sampling (Section~\ref{sec:headtohead}), since the only added work is candidate extraction at a single position.
For the trained baselines of Section~\ref{sec:headtohead}, rollout time per step in the same matched implementation is 1.34$\times$ for Token-MT (token-level entropy tracked over a 100K-token rolling window at every decoding step), 3.80$\times$ for Rollout-MT (longer high-temperature trajectories), and 4.81$\times$ for LATR (lookahead branches that may later be pruned). The max-entropy control additionally requires per-position entropy computation, which we did not profile. 
Concretely, for each batch of unique prompts, we 
(i) obtain the top-$N$ first-token candidates and their log-probabilities at the response-start position via a standard log-probability query, 
(ii) sample $K$ candidates without replacement on the host, 
(iii) append each selected token id to the prompt token ids of its assigned rollouts, and 
(iv) generate all $G$ continuations with the standard vLLM decoder~\citep{kwon2023efficient}.
No custom logits processor or separate per-branch decoding engine is required.
Prefix caching reuses shared prompt KV blocks, after which each branch maintains its own KV state exactly as in ordinary GRPO rollouts. 
Continuous batching and standard packing are preserved, and the rollout budget and active-sequence count are unchanged. 
The only extra work is candidate extraction and CPU-side assignment.

\subsection{Response-Start Context \texorpdfstring{$c(x)$}{c(x)} per Model}
\label{app:cx}
Table~\ref{tab:cx} instantiates the definition of Section~\ref{sec:observations} for every model in the paper, including the \thinktok-style scaffolds of recent families \citep{guo2025deepseek, qwen35blog, qwen3.6-27b}. To identify the first semantic token, we skip over the chat template's assistant prefix, structural delimiters, and whitespace-only tokens, and these preceding tokens remain part of the conditioning context $c(x)$.

\begin{table}[h]
\centering
\caption{Per-model instantiation of the response-start context.}
\label{tab:cx}
\small
\renewcommand{\arraystretch}{1.1}
\begin{tabular}{ll}
\toprule
Model & Scaffolding preceding the first token (included in \texorpdfstring{$c(x)$}{c(x)}) \\
\midrule
Qwen2.5-\{0.5B,3B,7B,14B,72B\}-Instruct & assistant prefix + \thinktok\texttt{\textbackslash n} \\
Llama3.2-3B-Instruct & assistant prefix + \thinktok\texttt{\textbackslash n} \\
Qwen3.5-9B / Qwen3.6-27B (non-thinking) & assistant prefix + empty \thinktok\texttt{</think>} pair \\ 
Gemma4-31B (base / it) & model-turn prefix (base: none) \\ 
DeepSeek-R1-Distill-Qwen-7B & assistant prefix + \thinktok\texttt{\textbackslash n} \\
Qwen2.5-Coder-3B-Instruct & assistant prefix \\
\bottomrule
\end{tabular}
\end{table}

\section{Semantic-Diversity Computation}
\label{app:semantic_diversity}

To test whether first-token choice affects the continuation rather than only the first word, we measure cosine semantic diversity after removing the first token itself. 
Each stripped continuation is embedded with \texttt{all-mpnet-base-v2}~\citep{song2020mpnet,reimers2019sentence}, following prior works~\citep{kirkunderstanding,karouzos2026does,zhao2026reinforced,bai2026learning} that compute semantic diversity using mean pairwise cosine distance between sentence embeddings.

For a rollout group $\mathcal{G}_x=\{y_i\}_{i=1}^{G}$, let $e(\operatorname{strip}(y_i))$ denote the embedding of the continuation. We define within-prompt semantic diversity as the average pairwise cosine distance:
\begin{equation}
D_{\mathrm{sem}}(\mathcal{G}_x)
\;=\;
\frac{2}{G(G-1)}
\sum_{1\le i<j\le G}
\Bigl[1-\cos\!\bigl(e(\operatorname{strip}(y_i)),\,e(\operatorname{strip}(y_j))\bigr)\Bigr].
\label{eq:semantic_diversity}
\end{equation}

We report the average of \(D_{\mathrm{sem}}(\mathcal{G}_x)\) over prompts.
Higher values indicate more semantically diverse continuations within the rollout group.

\section{Rollout Distribution and Training Loss}
\label{app:off-policy}

In standard rollout sampling, the first token after $c(x)$ is drawn from the behavior policy,
\[
f \sim \pi_{\theta_{\mathrm{old}}}(\cdot \mid c(x)).
\]
\Reft{} instead selects first tokens through its top-$N$ allocation rule, giving broader coverage to candidates that would rarely be sampled from the original first-token distribution.
Conditional on the selected first token, the continuation is generated from the same behavior policy as in the baseline,
\[
z \sim \pi_{\theta_{\mathrm{old}}}(\cdot \mid c(x),f).
\]
Thus, the distributional change enters through first-token selection, while the conditional continuation decoder remains unchanged.

We apply the same GRPO/DAPO loss to the resulting rollouts without an additional importance correction for the modified first-token sampling distribution.
This keeps the optimization procedure fixed while changing only how the rollout group is constructed.
Consequently, the update reflects \reft{}'s deliberate reallocation of the finite rollout budget across plausible first tokens, rather than reweighting the collected trajectories back toward the sharply concentrated original prior.
PPO-style clipping remains part of the underlying GRPO/DAPO objective and stabilizes updates between the current and behavior policies.
For lower-ranked candidates, whose policy probability is orders of magnitude below $1/N$, an explicit importance correction would strongly downweight rollouts starting from those first tokens and pull the update back toward the concentrated prior that \Reft{} is designed to broaden.

This setup is related in spirit to RLVR interventions that modify rollout collection or conditioning while keeping the optimization fixed, such as temperature-scheduled generation \citep{zhuang2025exploring, yang2025let} and prefix-conditioned or prefix-reuse methods \citep{setlur2026reuse, huang2026pros, li2025staying, kim2026training}.

Additionally, monitoring training throughout on Qwen2.5-3B-Instruct and GSM8K, reward curves are smooth under both methods. 
The fraction of policy-loss tokens outside the clipping range stays below 0.02\% on average for both (means $8.18\times10^{-5}$ for GRPO and $1.61\times10^{-4}$ for \reft; maxima $1.51\times10^{-3}$ and $2.14\times10^{-3}$), and the mean gradient norm is higher under \reft{} (0.049 vs.\ 0.036), with no gradient spikes, reward collapse, or degenerate generation.
Apart from the first-token sampler, the answer verifier, reward function, advantage computation, RL objective, model architecture, continuation decoder, and total rollout budget are otherwise unchanged.
Across the experiments in Section~\ref{sec:experiments}, this modified rollout distribution trains stably and yields the performance gains reported in the main results.

Appendix~\ref{app:utility} further evaluates general-purpose capabilities under standard inference and finds no noticeable degradation relative to GRPO on instruction following, general reasoning, and knowledge benchmarks.

\newpage

\section{Additional Analysis}
\label{app:further_anlaysis}

\begin{wrapfigure}{r}{0.5\linewidth}
\vspace{-14pt}
\centering
\includegraphics[width=\linewidth]{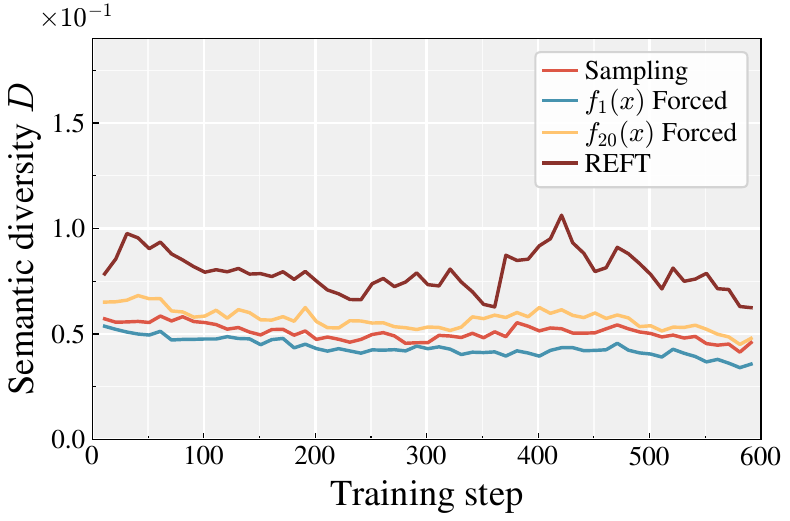}
\vspace{-6pt}
\caption{
\textbf{\Reft{} broadens exploration throughout training.}
Semantic diversity $D$ of rollout groups collected during training, measured as the dispersion of continuation embeddings within each group.
\Reft{} stays above standard sampling and the fixed first-token controls $f_1(x)$ and $f_{20}(x)$ at every step, and the gap persists as all curves gradually decline.
}
\label{fig:sem_div_training}
\vspace{10pt}
\end{wrapfigure}
\paragraph{Training-time diversity.}
Figure~\ref{fig:sem_div_training} tracks the semantic diversity of the rollout groups each method collects during training, measured on continuation embeddings.
\Reft{} yields markedly more diverse groups than standard sampling throughout training.
Diversity drifts downward for every method as the policy sharpens, as expected under reward-driven optimization, but \Reft{} remains above the alternatives at every step rather than converging to them.
The fixed-token controls show that this gain is not a generic effect of first-token forcing.
Fixing $f_{20}(x)$ roughly tracks standard sampling and fixing $f_1(x)$ falls below it, so forcing any single token does not by itself widen the group.
Because \Reft{} changes only how rollouts are drawn, the added diversity reflects the broader set of alternatives the policy is exposed to under group-relative updates, while the objective itself still rewards verifier success rather than output dispersion.
Read alongside Tables~\ref{tab:main_pass1_pass8} and~\ref{tab:pass64}, these results indicate that sustained exploration during training accompanies higher single-sample accuracy and larger-budget success rates.

\begin{wrapfigure}{r}{0.5\textwidth}
    \centering
    \vspace{-1.5em}
    \includegraphics[width=\linewidth]{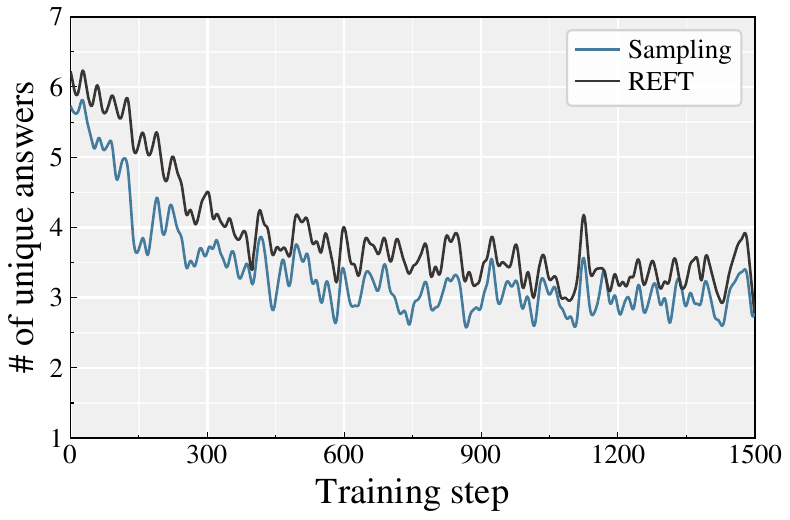}
    \vspace{-1.2em}
    \caption{
    \small
    \textbf{Within-group answer diversity.}
    Number of unique final answers per rollout group during training on Qwen2.5-0.5B-Instruct, GSM8K, DAPO.
    }
    \label{fig:num_of_answer_correct}
    \vspace{-1.3em}
\end{wrapfigure}

\paragraph{Outcome diversity beyond trajectory diversity.}
The near-flat first-token-conditioned accuracy in Section~\ref{sec:observations} shows viability, not equivalence.
Plausible alternative first tokens are not destructive to correctness on average, but they can still route generation into different continuation regions.
We therefore complement trajectory-level semantic diversity with final-answer diversity, a tractable proxy for outcome-space coverage used in prior work on reasoning diversity~\citep{song2025outcome,dang2025assessing}.
Figure~\ref{fig:num_of_answer_correct} shows that \Reft{} produces more distinct answers than standard sampling during training, suggesting that first token diversification changes not only the reasoning text but also the set of solution attempts under the same rollout budget.
More distinct answers alone do not establish useful exploration.
This diagnostic complements the analysis in Section~\ref{sec:analysis}, where the zero-variance decomposition
in Figure~\ref{fig:zero_var} shows fewer all-wrong zero-variance
groups under \Reft{} in reported comparisons.

\newpage

\section{Qualitative Math Rollout Examples}
\label{app:qualitative}

We present two logged GSM8K rollout examples illustrating the rollout-level effect of \Reft{}.
Both examples are sampled from Qwen2.5-3B-Instruct trained with GRPO on GSM8K.
For each prompt, we contrast eight standard-sampling rollouts with eight \Reft{} rollouts produced as $K{=}4$ first-token branches with $G/K{=}2$ continuations per branch.
We display representative continuations from each condition rather than all sixteen completions.
 
\subsection{Case 1: Water Consumption Problem (gold answer = 26)}
\label{app:qualitative_q102}
 
\begin{promptbox2}[title=Prompt]
John likes to have a glass of water with breakfast, lunch and dinner. Finally, he has one before he goes to bed as well. John does this every weekday, but on the weekends he likes to relax and have a soda with dinner instead. How many glasses of water does John drink in a week?
\end{promptbox2}

\paragraph{Standard sampling: $0/8$ correct.}
The rollouts open almost exclusively with \texttt{To} or \texttt{First}, and all eight reach incorrect answers ($7, 15, 17, 17, 20, 22, 22, 22$).
They make several distinct bookkeeping errors.
Some, like rollout~1 below, undercount the daily water occasions, while others, like rollout~6, identify the four weekday occasions correctly but undercount weekend consumption after replacing dinner with soda.

\begin{rolloutbox}[title={Standard rollout 1}]
\texttt{<think>}\\
To determine the total number of glasses of water John drinks in a week, we need to break down his water consumption pattern and calculate accordingly.
 
John drinks one glass of water with each meal throughout the week:
- For breakfast \;-\; For lunch \;-\; For dinner
 
\textit{[\dots]}
 
On weekends (Saturday and Sunday), he drinks a soda instead of water during dinner:
- Saturday: 1 glass of water \;-\; Sunday: 1 glass of water
 
$5\text{ (weekdays)} + 2\text{ (weekend days)} = 7\text{ glasses}$\\
\texttt{</think>} \quad \texttt{<answer>} 7 \texttt{</answer>}
\end{rolloutbox}
 
\begin{rolloutbox}[title={Standard rollout 6}]
\texttt{<think>}\\
To determine how many glasses of water John drinks in a week, we need to break down his drinking habits by weekdays and weekends.
 
1. \textbf{Weekdays}: John drinks water with breakfast, lunch, dinner, and before bed. That gives $4$ glasses per weekday $\times 5$ weekdays $= 20$ glasses.
 
\textit{[\dots]}
 
2. \textbf{Weekends}: He has a soda with dinner on weekend days, but still drinks water at the other meals.
 
$20 + 2 = 22\text{ glasses}$\\
\texttt{</think>} \quad \texttt{<answer>} 22 \texttt{</answer>}
\end{rolloutbox}
 
\paragraph{\Reft{}: $8/8$ correct.}
\Reft{} allocates its eight rollouts across four selected first-token branches, \texttt{Let}, \texttt{First}, \texttt{Given}, and \texttt{Initially}, with two continuations each.
Each opens a different framing of the same problem (step-by-step decomposition, weekday-first, premise-restatement, sequential narration), and all eight continuations reach the correct answer $26$.

\begin{rolloutbox}[title={\Reft{} rollout 1 \quad First Token = \texttt{Let}}]
\texttt{<think>}\\
Let's break down the problem step by step.
 
1. John drinks water at the following times: breakfast, lunch, dinner, and before bed.
2. Therefore, John drinks water $4$ times a day.
3. There are $5$ weekdays and $2$ weekend days.
4. On weekdays: $5\times 4 = 20$ glasses.
5. On weekends, he switches from water to soda for dinner, so he drinks $3$ glasses each weekend day, $2\times 3 = 6$ glasses.
 
$20 + 6 = 26$ glasses.\\
\texttt{</think>} \quad \texttt{<answer>} 26 \texttt{</answer>}
\end{rolloutbox}
 
\begin{rolloutbox}[title={\Reft{} rollout 5 \quad First Token = \texttt{Given}}]
\texttt{<think>}\\
Given the information, John drinks water with his meals every weekday and a soda with his dinner on weekends. A week consists of 5 weekdays and 2 weekend days.
 
\textit{[\dots]}
 
Weekdays contribute $5 \times 4 = 20$ glasses, and weekend days contribute $2 \times 3 = 6$ glasses, for a total of $26$ glasses.\\
\texttt{</think>} \quad \texttt{<answer>} 26 \texttt{</answer>}
\end{rolloutbox}
 
\begin{rolloutbox}[title={\Reft{} rollout 7 \quad First Token = \texttt{Initially}}]
\texttt{<think>}\\
Initially, let's determine how many glasses of water John drinks on a regular weekday. According to the problem, John drinks one glass at breakfast, lunch, dinner, and before bed, giving $4$ glasses per weekday.
 
\textit{[\dots]}
 
For weekends, dinner is replaced by soda, leaving $3$ glasses of water per day. Total: $5\times 4 + 2\times 3 = 26$ glasses.\\
\texttt{</think>} \quad \texttt{<answer>} 26 \texttt{</answer>}
\end{rolloutbox}

\paragraph{Why this case is informative.}
Standard sampling does not fail by collapsing onto a single wrong answer.
Its rollouts make several distinct bookkeeping errors, and none keeps the weekday and weekend accounting consistent at the same time.
\Reft{}'s alternative first-token conditions more often route into decompositions that keep this accounting consistent, and here all eight continuations reach the correct total, with each displayed rollout separating the four weekday occasions from the three weekend ones.
Since everything after the opener is sampled with the same decoder under both conditions, this outcome is consistent with the routing effect described in Section~\ref{sec:observations}.

\subsection{Case 2: Lemon Tree Problem (gold answer = 13)}
\label{app:qualitative_q012}
 
\begin{promptbox2}[title=Prompt]
Carlos is planting a lemon tree. The tree will cost \$90 to plant. Each year it will grow $7$ lemons, which he can sell for \$1.5 each. It costs \$3 a year to water and feed the tree. How many years will it take before he starts earning money on the lemon tree?
\end{promptbox2}

\paragraph{Standard sampling: $0/8$ correct.}
Five of the eight rollouts return the same wrong answer of $12$ (computing the year in which cumulative profit is non-negative, rather than the year in which it is strictly positive); the remaining three return $9$, $73.5$, and $1$.
The five rollouts that answer $12$ differ in surface text but share the same off-by-one error.

\begin{rolloutbox}[title={Standard rollout 1}]
\texttt{<think>}\\
To determine how many years it will take before Carlos starts earning money on his lemon tree, we need to set up an inequality based on his profit.
 
\textit{[\dots]}
 
$7.5y \geq 90 \;\Rightarrow\; y \geq 12$
 
So Carlos will start earning money in year $12$.\\
\texttt{</think>} \quad \texttt{<answer>} 12 \texttt{</answer>}
\end{rolloutbox}
 
\paragraph{\Reft{}: $7/8$ correct.}
\Reft{} allocates its eight rollouts across four selected first-token branches, \texttt{First}, \texttt{Carlos}, \texttt{To}, and \texttt{When}, with two continuations each.
Seven of the eight rollouts correctly compute that the tree starts \emph{strictly earning} money in year $13$.
 
\begin{rolloutbox}[title={\Reft{} rollout 1 \quad First Token = \texttt{First}}]
\texttt{<think>}\\
First, we need to determine if the lemon tree will generate more money through sales than it costs to maintain.
 
\textit{[\dots]}
 
Annual net profit: $7 \times 1.5 - 3 = 7.5$. We need $7.5 y > 90$, i.e., $y > 12$. The smallest such integer is $y = 13$.\\
\texttt{</think>} \quad \texttt{<answer>} 13 \texttt{</answer>}
\end{rolloutbox}
 
\begin{rolloutbox}[title={\Reft{} rollout 3 \quad First Token = \texttt{Carlos}}]
\texttt{<think>}\\
Carlos starts by investing \$90 to plant the lemon tree. Each year, he has to spend \$3 to water and feed it, while earning $7 \times \$1.5 = \$10.5$ from selling lemons.
 
\textit{[\dots]}
 
Cumulative net profit after $y$ years is $7.5y - 90$, which becomes positive at $y = 13$.\\
\texttt{</think>} \quad \texttt{<answer>} 13 \texttt{</answer>}
\end{rolloutbox}
 
\begin{rolloutbox}[title={\Reft{} rollout 7 \quad First Token = \texttt{When}}]
\texttt{<think>}\\
When calculating the number of years it will take before Carlos starts earning money on the lemon tree, we need to consider both the initial cost and the recurring annual costs and income.
 
\textit{[\dots]}
 
The break-even is at $y = 12$, so Carlos first \emph{earns} money in year $13$.\\
\texttt{</think>} \quad \texttt{<answer>} 13 \texttt{</answer>}
\end{rolloutbox}
 
\paragraph{Takeaway.}
Across both cases, the difference between standard sampling and \Reft{} is not that \Reft{} produces stranger or longer text.
The completions remain coherent English with valid \texttt{<think>}/\texttt{<answer>} structure and no degenerate behavior.
What changes is that diversifying the first token spreads the group over several framings of the same problem, and these framings can differ in how they set up the computation and where they conclude.
These examples illustrate at the rollout level the effect that the aggregate analyses in Section~\ref{sec:analysis} measure.


\section{Qualitative Code Rollout Examples}
\label{app:qualitative_code}

We provide two logged examples from Qwen2.5-Coder-3B-Instruct,
where generation begins directly with code rather than an explicit reasoning trace.
As in the mathematical examples, each prompt has $G{=}8$ rollouts. 
Under \reft{}, $K{=}4$ first-token branches receive two continuations each.
We omit fixed structural prefixes such as code fences and template
comments and display each rollout beginning at its first semantic
code token.
Only representative beginnings are shown, since the purpose is to
illustrate how the first-token choice shapes the subsequent program.

\subsection{Case 1: Common Substring Detection}
\label{app:qualitative_code_twostrings}

\begin{promptbox2}[title=Prompt (abridged)]
Given two strings $s_1$ and $s_2$, determine whether they share
a common substring.
A substring may consist of a single character.
For each of $p$ test cases, output \texttt{YES} if the two strings
share at least one character and \texttt{NO} otherwise.
\end{promptbox2}

\paragraph{Standard sampling: $1/8$ full-test pass.}
Only one of the eight standard rollouts receives full sandbox-test
reward.
Several failures reach the correct set-intersection idea but break
elsewhere in the surrounding program.
For example, rollout 3 below replaces input with sys.stdin.read and then attempts to parse the entire input as the test-case count.
 
\begin{rolloutbox}[title=Standard rollout 2]
\begin{Verbatim}
def twoStrings(s1, s2):
    set1 = set(s1)
    set2 = set(s2)
    if set1 & set2:
        return "YES"
    else:
        return "NO"
 
def main():
    import sys
    input = sys.stdin.read
    data = input().split()
    p = int(data[0])
\end{Verbatim}
\textit{[\dots]}
\end{rolloutbox}
 
\begin{rolloutbox}[title=Standard rollout 3]
\begin{Verbatim}
def twoStrings(s1, s2):
    set1 = set(s1)
    set2 = set(s2)
    if set1.intersection(set2):
        return "YES"
    else:
        return "NO"
 
import sys
input = sys.stdin.read
p = int(input().strip())
\end{Verbatim}
\textit{[\dots]}
\end{rolloutbox}
 
\paragraph{\Reft{}: $6/8$ full-test pass.}
The four selected first semantic tokens are
\texttt{for}, \texttt{s}, \texttt{input}, and \texttt{T},
at policy ranks 5, 7, 8, and 11, respectively.
Six of the eight resulting continuations receive full reward.
The successful branches organize the same underlying computation
differently from the first line onward.
 
\begin{rolloutbox}[title={\Reft{} rollout 1 \quad First Token = \texttt{for}}]
\begin{Verbatim}
for _ in range(int(input())):
    s1 = input()
    s2 = input()
    if len(set(s1) & set(s2)) > 0:
        print("YES")
    else:
        print("NO")
\end{Verbatim}
\end{rolloutbox}
 
\begin{rolloutbox}[title={\Reft{} rollout 7 \quad First Token = \texttt{T}}]
\begin{Verbatim}
T = int(input().strip())
for _ in range(T):
    s1 = input().strip()
    s2 = input().strip()
    if any(c in s2 for c in s1):
        print("YES")
    else:
        print("NO")
\end{Verbatim}
\end{rolloutbox}
 
The successful branches differ in their input handling and top-level
control flow despite originating from extremely low-probability
first tokens.
Thus, changing the first semantic token affects more than its lexical
identity. 
It changes how the continuation organizes the program.

\subsection{Case 2: Counting Zeros in a Binary Representation}
\label{app:qualitative_code_binary}

\begin{promptbox2}[title=Prompt (abridged)]
For each input integer $N$, output the number of zeros in its binary
representation.
For example, $2=10_2$ contains one zero and $4=100_2$ contains two.
\end{promptbox2}

\paragraph{Standard sampling: $0/8$ full-test pass.}
None of the eight standard rollouts receives full sandbox-test reward.
A representative failure applies the right high-level operation but
counts zeros directly in Python's \texttt{bin()} representation,
thereby also counting the zero in its \texttt{0b} prefix.
 
\begin{rolloutbox}[title=Standard rollout 3]
\begin{Verbatim}
def count_zeros_in_binary(n):
    return bin(n).count('0')
 
def main():
    import sys
    T = int(sys.stdin.readline())
    for _ in range(T):
        N = int(sys.stdin.readline())
        print(count_zeros_in_binary(N))
\end{Verbatim}
\textit{[\dots]}
\end{rolloutbox}
 
\paragraph{\Reft{}: $5/8$ full-test pass.}
The four selected first semantic tokens are
\texttt{from}, \texttt{test}, \texttt{input}, and \texttt{number},
at policy ranks 3, 11, 13, and 14, respectively.
Five of the eight continuations receive full reward.
Successful branches recover multiple implementations rather than
collapsing onto a single code template.
 
\begin{rolloutbox}[title={\Reft{} rollout 2 \quad First Token = \texttt{from}}]
\begin{Verbatim}
from sys import stdin
 
def count_zeros(n):
    count = 0
    while n > 0:
        if n % 2 == 0:
            count += 1
        n //= 2
    return count
\end{Verbatim}
\textit{[\dots]}
\end{rolloutbox}
 
\begin{rolloutbox}[title={\Reft{} rollout 3 \quad First Token = \texttt{test}}]
\begin{Verbatim}
test_cases = int(input())
for test in range(test_cases):
    N = int(input())
    binary_representation = bin(N)[2:]
    count_of_zeros = binary_representation.count('0')
    print(count_of_zeros)
\end{Verbatim}
\end{rolloutbox}
 
\paragraph{The first token does not determine the algorithm.}
The two successful branches above develop distinct implementations:
one counts zero bits through repeated arithmetic operations, while the
other converts the number to a binary string and removes the
\texttt{0b} prefix before counting.
Moreover, even two continuations conditioned on the same
\texttt{from} first token can differ in correctness.
The first token therefore acts as an early routing choice rather than
a complete commitment to a particular implementation.

\paragraph{Takeaway.}
Across both examples, \reft{} introduces variation before the program
has committed to its later syntax, control flow, and implementation
details.
Low-probability first tokens can seed coherent and successful program
trajectories that standard sampling would rarely visit, while all
subsequent tokens are generated normally by the policy.
In this sense, first-token diversification expands the set of program
routes exposed to training without repeatedly perturbing an already
formed program.



\end{document}